\documentclass[sigconf]{acmart}

\usepackage{amsthm}
\usepackage{algorithm}
\usepackage{algorithmic}
\newlength\myindent
\newcommand\bindent{%
  \begingroup
  \setlength{\itemindent}{\myindent}
  \addtolength{\algorithmicindent}{\myindent}
}
\newcommand\eindent{\endgroup}

\AtBeginDocument{%
  \providecommand\BibTeX{{%
    \normalfont B\kern-0.5em{\scshape i\kern-0.25em b}\kern-0.8em\TeX}}}

\copyrightyear{2023}
\acmYear{2023}
\setcopyright{acmlicensed}\acmConference[GECCO '23]{Genetic and Evolutionary Computation Conference}{July 15--19, 2023}{Lisbon, Portugal}
\acmBooktitle{Genetic and Evolutionary Computation Conference (GECCO '23), July 15--19, 2023, Lisbon, Portugal}
\acmPrice{15.00}
\acmDOI{10.1145/3583131.3590508}
\acmISBN{979-8-4007-0119-1/23/07}

\begin{document}

%% allowing the author to define a "short title" to be shown in page headers.
\title{Accelerating Evolution Through Gene Masking\\and Distributed Search}

%% Of note is the shared affiliation of the first two authors and the
\author{Hormoz Shahrzad}
\affiliation{
\institution{Cognizant AI Labs \& UT Austin}
\city{San Francisco}
\country{USA}
}
\email{hormoz@cognizant.com}

\author{Risto Miikkulainen}
\affiliation{
\institution{UT Austin \& Cognizant AI Labs}
\city{Austin \& San Francisco}
\country{USA}
}
\email{risto@cognizant.com}

%%
%% By default, the full list of authors will be used on the page
%% headers. Often, this list is too long and will overlap
% \renewcommand{\shortauthors}{Shahrzad, et al.}

%%
%% The abstract summarizes the work to be presented in the article.
\begin{abstract}
In building practical applications of evolutionary computation (EC), two optimizations are essential. First, the parameters of the search method need to be tuned to the domain in order to balance exploration and exploitation effectively. Second, the search method needs to be distributed to take advantage of parallel computing resources. This paper presents BLADE (BLAnket Distributed Evolution) as an approach to achieving both goals simultaneously. BLADE uses blankets (i.e., masks on the genetic representation) to tune the evolutionary operators during the search, and implements the search through hub-and-spoke distribution. In the paper, (1) the blanket method is formalized for the (1 + 1)EA case as a Markov chain process. Its effectiveness is then demonstrated by analyzing dominant and subdominant eigenvalues of stochastic matrices, suggesting a generalizable theory; (2) the fitness-level theory is used to analyze the distribution method; and (3) these insights are verified experimentally on three benchmark problems, showing that both blankets and distribution lead to accelerated evolution. Moreover, a surprising synergy emerges between them: When combined with distribution, the blanket approach achieves more than $n$-fold speedup with $n$ clients in some cases. The work thus highlights the importance and potential of optimizing evolutionary computation in practical applications.
\end{abstract}

\begin{CCSXML}
<ccs2012>
   <concept>
       <concept_id>10010147.10010257.10010293.10011809.10011812</concept_id>
       <concept_desc>Computing methodologies~Genetic algorithms</concept_desc>
       <concept_significance>500</concept_significance>
       </concept>
   <concept>
       <concept_id>10003752.10003809.10003716.10011136.10011797.10011799</concept_id>
       <concept_desc>Theory of computation~Evolutionary algorithms</concept_desc>
       <concept_significance>500</concept_significance>
       </concept>
    <concept>
       <concept_id>10010147.10010919.10010172</concept_id>
       <concept_desc>Computing methodologies~Distributed algorithms</concept_desc>
       <concept_significance>300</concept_significance>
       </concept>
 </ccs2012>
\end{CCSXML}

\ccsdesc[500]{Computing methodologies~Genetic algorithms}
\ccsdesc[500]{Theory of computation~Evolutionary algorithms}
\ccsdesc[300]{Computing methodologies~Distributed algorithms}
\keywords{Evolutionary algorithms, genetic algorithms, distributed evolution, adaptive evolution, fitness-level method, Markov chains, stochastic matrices}

\nocite{miikkulainen:agebook18,moriarty:ec97,gerules:cec16,gomez:jmlr08,potter:ec00,Doerr_onemax:gecco2019,Doerr_onemax:gecco2020,doerr-papa:gecco2020,berman1994nonnegative,Kirkland2009SubdominantEF,Hassanat2019,Doerr_Benchmarking_2019,Lssig2014GeneralUB,Rowe2012TheCO,island,Multimodal2019,Strasser2017FactoredEA,ecstar2013}

\maketitle

\section{Introduction}
% In evolutionary computation, masking refers to hiding certain parts of a candidate solution by setting certain bits or genes in it to be "masked", which means they are not eligible to be modified by the evolutionary operators.

Automatic configuration, often called AutoML or meta-learning, has recently emerged as an important topic in machine learning \cite{elsken:jmlr19,liang:gecco19,liang:gecco21}. Complex machine learning systems depend on several hyperparameters that are difficult to set right by hand, and therefore machine learning itself is harnessed to optimize them.

Likewise, the efficacy of an evolutionary computation approach often relies on the appropriate calibration of its operators. This paper proposes a new method for doing so automatically as part of the evolutionary search itself. The idea is to use a mask, i.e.\ a blanket, on the genotype to focus the search on specific parts of the problem. The masks are constructed dynamically throughout the search, and help focus it on parts that are the most important. In a sense, they play a role similar to attention heads in transformer neural networks \cite{vaswani:neurips17}, but adapted to population-based search.

A related practical challenge in machine learning is parallelization. As machine learning systems grow in size, there is a growing need for distributing the computation across multiple computing resources, including multiple cores on a single machine, multiple nodes in a cluster, and multiple resources in the cloud. While evolutionary computation generally parallelizes well, the method of distribution of evaluations and evolutionary operations has a large effect. The hub-and-spoke method is often preferable for maximum scalability and flexibility.

Putting blankets and distribution together, this paper proposes BLADE as an effective new method for accelerated evolution. Each of its components is first formalized and characterized theoretically, leading to predictions of possible speedups. These predictions are then confirmed in practical experiments with the $(1+1)EA$ optimization method on three optimization benchmarks, i.e.\  AllOnes, OneMax, and LeadingOnes. The results reveal a surprising synergy between blankets and distribution that allows more than $n$-fold speedups with $n$ clients. In future work, the BLADE approach may be extended to other algorithms and applications, and can thus serve as a foundation for accelerated evolution in practice.

\section{Background}

This section reviews prior work related to BLADE both in automatic parameter tuning and in distributed evolutionary computation.

\subsection{Parameter Optimization}
\label{sc:ParameterOptimization}

All problem-solving methods rely on a number of parameters that have to be set appropriately for the method to function properly. In genetic algorithms, they include settings for the population size, the mutation rate and extent, the type of crossover, the selection method, the size of the elite set, the number of offspring, etc.  They can be set up by hand through a laborious trial and error process, or a learning method such as evolution itself can be used to discover good settings.

For instance, in bilevel evolution, low-level evolution searches for solutions while high-level evolution searches for the best parameters for low-level evolution \cite{sinha:gecco14,liang:gecco15}. Bilevel evolution is expensive because it often requires running many low-level optimizations to evaluate the fitness of high-level individuals. Furthermore, there is a growing body of evidence suggesting that operator settings and other aspects of the configurations should be adapted dynamically in response to changes in the fitness of the population \cite{autoconfig2022gecco,ansotegui2009gender,AdaptingOperatorSettings,Hassanat2019,doerr-papa:gecco2020,Doerr_Benchmarking_2019,Doerr_onemax:gecco2019}.

The blanket method establishes such a dynamic optimization mechanism: The settings of the search operator are adjusted as part of the search itself over the course of the run.

\subsection{Distributed Evolution}
\label{sc:DistributedEvolution}

There are several different methods for distributing an evolutionary process across multiple clients, and each method has its own advantages and disadvantages.

\emph{Synchronous distribution} \cite{ParallelEA2015,parallel2019,ParallelAD2021,Hassani2009AnOO}: The evolution engine partitions the population, assigns each partition to an external worker to evaluate, and waits for all the evaluations to return before forming the population for the next generation. This method is simple, but it only scales well when the worker clients are machines with similar speed and connectivity; otherwise, time is wasted waiting for the slowest clients to finish their work. This approach can also be slow for large populations because there is a high communication overhead for distributing individuals.

\emph{Asynchronous population evaluation} \cite{ParallelEA2015,parallel2019,ParallelAD2021,Hassani2009AnOO} starts like the synchronous method, but the engine only waits until a part of the population has been evaluated before generating the next population. This method improves efficiency but may lose diversity.

\emph{Island model} \cite{island}: While the above methods rely on the star topology of distribution, the island model applies to a variety of topologies such as a ring or hypercube. The evolution engines themselves are distributed, and they migrate good solution candidates between themselves using peer-to-peer connections. This method is asynchronous and can generate a lot of diversity at scale. On the other hand, decentralization of evolution engines makes harvesting the best candidates difficult, and the peer-to-peer connections may become an overhead, especially when using highly connected topologies on different physical machines. Therefore, the island model is most suitable for parallel evolution on a single machine with many cores rather than distributing it over a network of machines.

\emph{Hub and spoke model} \cite{ecstar2013}: This model also uses the star topology, but the clients (spokes) are full evolution engines. They evaluate candidates asynchronously on different partitions of the data and the results are aggregated in a centralized server (hub). The main advantage of this method is that it can be easily distributed over a very large network of diverse computing resources. Also, each client has only one external connection to the hub, and therefore it is possible to add or remove clients in an asynchronous manner.  On the other hand, the hub may become overwhelmed with a very large number of clients. In such a situation, a method of load-balancing can be implemented.

In sum, the choice of distribution method depends on the specific requirements of the evolutionary algorithm and the computational resources available. For many evolutionary computation implementations, the hub-and-spoke model offers the best advantages, and will thus be used for BLADE as well.

\section{Method}
\label{sc:method}

This section presents the details of the BLADE method. Intuition and theoretical formulation are first given for blankets and distribution separately, and these methods are then brought together to full BLADE. The discussion focuses on the optimization of binary strings with $(1+1)EA$. Possibilities for extending BLADE to other representations and search methods will be discussed in Section~\ref{sc:discussion}.

\subsection{Blanket-based Search}
\label{sc:blanket-search}

The blanket method refers to the process of masking parts of a candidate solution so that they cannot be modified by the evolutionary operators such as mutation for $(1+1)EA$. It is a way to focus the search on those solution elements that make progress most likely.

\subsubsection{Method Description}

The blanket method involves modifying the mutation rate of bits in a binary string of length $N$ according to another binary string, i.e.\ the blanket, which can be any one of the $2^N$ such strings except all zeros and all ones. The mutation rate $\mu$ is modified by the factor of $\frac{N}{N-{\mathrm{len(blanket)}}}$.

For example, if $N=5$, $\mu = 1/4$, and the blanket is "$01011$", then $\mathrm{len(blanket)} = 3$, and the second, fourth, and fifth bits are preserved, and the first and third bits are mutated independently with the modified mutation probability $\mu_\mathrm{b}=\frac{1}{4}*\frac{5}{5-3}=\frac{5}{8}$. If $\mu_\mathrm{b}$  exceeds 1.0, it is clipped to 1.0, meaning that all bits not under the blanket are flipped deterministically. Algorithm \ref{alg:blanket} shows a modified version of $(1+1)EA$ that incorporates the blanket method.

\begin{algorithm}[ht]
   \caption{Blankets only (non-distributed BLADE)}
   \label{alg:blanket}
   \begin{algorithmic}
   \STATE {\bfseries Initialization:}
   \bindent
   \STATE Sample $x \in \{0,1\}^N$ uniformly at random and evaluate $f(x)$
   \eindent
   \STATE {\bfseries Optimization:}
   \bindent
   \FOR{$t=1,2,3,...$}
   \STATE {\bfseries Offspring generation with blanket:} 
   \bindent
   \STATE Calculate mutation rate $\mu$
   \STATE Sample \emph{blanketLength} from $[1, N-1]$ randomly
   \STATE Set \emph{blanket} $\gets$ flip \emph{blanketLength} random bits in $\{0\}^N$
   \STATE Set $\mu_\mathrm{b} \gets \min(1, \mu(\frac{N}{N-\emph{blanketLength}}))$
   \STATE Sample $y \in \{0,1\}^N$ at random with $p(1)=\mu_\mathrm{b}$
   \STATE Set $\emph{blanket} \gets \emph{blanket} \land y$ ($\land$ is bitwise AND)
   \STATE Create $x^* \gets x \oplus$ \emph{blanket} ($\oplus$ is bitwise XOR)
   \eindent
   \STATE {\bfseries Selection:} 
   \bindent
   \IF{$f(x^*) \geq f(x)$} 
   \STATE $x \gets x^*$ 
   \ENDIF
   \eindent
   \ENDFOR
   \eindent
   \end{algorithmic}
\end{algorithm}

\subsubsection{Theoretical Formulation}

\begin{figure*}[ht]
  \centering
  \hfill
  \begin{minipage}[b]{0.48\linewidth}
  \centering
  \includegraphics[width=\linewidth]{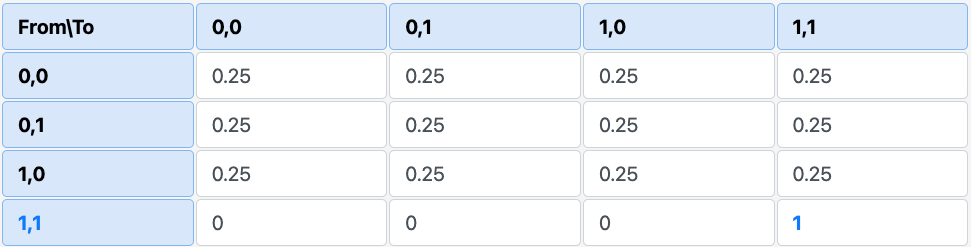}\\
  (a) Example transition matrix without blankets
  \end{minipage}
  \hfill
  \begin{minipage}[b]{0.48\linewidth}
  \centering
  \includegraphics[width=\linewidth]{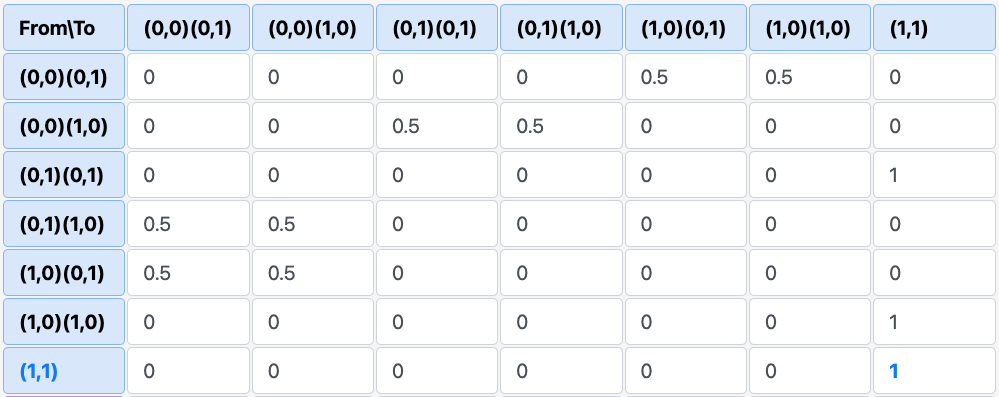}\\
  (b) Example transition matrix with blankets
  \end{minipage}\\[2ex]
  \hfill
  \begin{minipage}{0.40\linewidth}
  \centering
  \includegraphics[width=\linewidth]{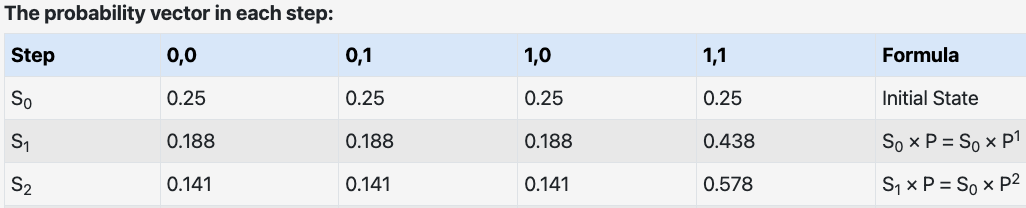}\\
  \vspace*{-1ex}
  $\dots$\\[1ex]
  \includegraphics[width=\linewidth]{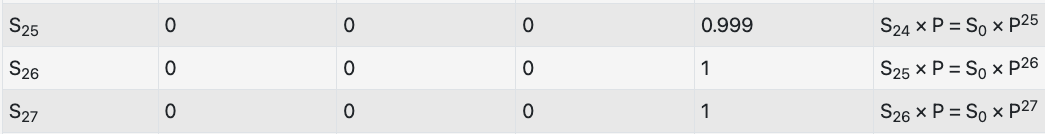}\\
  (c) Convergence without blankets
  \end{minipage}
  \hfill
  \begin{minipage}{0.48\linewidth}
  \centering
  \includegraphics[width=\linewidth]{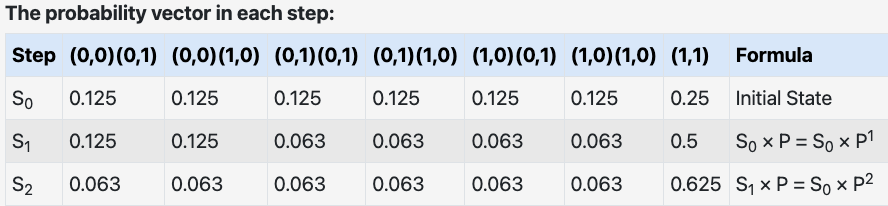}\\
  \vspace*{-1ex}
  $\dots$\\[1ex]
  \includegraphics[width=\linewidth]{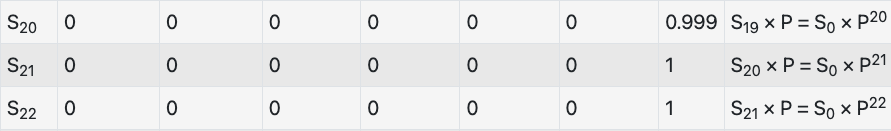}\\
  (d) Convergence with blankets
  \end{minipage}
\caption{A comparison of convergence with and without blankets on a 2-bit AllOnes problem is shown using transition matrices (a) and (b). The presence of two possible blankets for each state in (a) results in twice as many states in (a), except for the last converging state of (1,1). The subdominant eigenvalue is 0.75 for matrix (a) and 0.71 for matrix (b), indicating faster theoretical convergence with blankets. The convergence process is depicted in (c) and (d), starting from a randomly chosen initial state $S_0$ with uniform probability. The process takes 26 transitions to converge without blankets, and only 21 transitions with blankets, confirming the theoretical advantage of using blankets in the search.}
  \label{fig:markov}
\end{figure*}

The blanket method may initially seem counterintuitive: It may help search if it masks those parts of the string that are indeed part of the solution, but it may also hinder search if it masks parts where further mutations are needed. Without further knowledge of the problem domain or a method for dynamically constructing blankets, its potential benefits and drawbacks may cancel out. A simple formalization is helpful in showing why this is not a problem, and blankets can indeed speed up the search process.

Consider $(1+1)EA$ on a 2-bit version of the AllOnes problem: Starting from random bits, each bit gets independently mutated at each iteration with the probability of $\frac{1}{2}$ until both bits are one. This process can be formalized as a Markov chain, i.e.\ a random process in which the future state of the system depends only on its current state and not its past history \cite{MarkovCF}. A transition matrix stores the probabilities of the system moving from one state to another, with zero indicating that a transition is not possible. An absorbing state is a state that cannot be left once it has been entered; it is represented by a row with a single non-zero element of 1.

Since there are $2^2=4$ different combinations of two bits, the transition matrix is of size $4 \times 4$. A possible such matrix is shown in Figure~\ref{fig:markov}(a). The rows represent the current states, the columns the possible next states, and each cell the probability of the corresponding transition. Once the system gets to the state in the bottom row it stays there forever; otherwise, it can transition from any state to any other state with a probability of 0.25.

With two bits, there are two possible blankets, "[0,1]" and "[1,0]". Thus the number of states doubles, except the absorbing state in which the blanket has no effect, so the transition matrix expands to $7 \times 7$, as shown in Figure~\ref{fig:markov}(b). For instance, from the state (0,0) with the blanket of (0,1), i.e.\ the top left cell, the system cannot go to state (0,0): By masking the second bit, the $\frac{1}{2}$ baseline probability of mutation increases by the factor of $\frac{N}{N-\mathrm{len(blanket)}}=\frac{2}{2-1}=2$ and thus becomes 1. Therefore, the system can only transition to the state (1,0); since there are two possibilities for the blankets in that state, their probabilities are 0.5 for both.

Note that an $n$ × $n$ entrywise nonnegative matrix $P$ is considered to be \textit{stochastic} if the sum of its every row is equal to 1, i.e.\ $P\textbf{1} = \textbf{1}$, where $\textbf{1}$ is an all-ones column vector of size $n$. Transition matrices such as those in Figure~\ref{fig:markov} above are thus stochastic matrices. Further, from the definition of eigenvalue and eigenvector (i.e., $P\textbf{x}=\lambda \textbf{x}$), it is clear that 1 is an eigenvalue of $P$. According to the Perron-Frobenius theorem \cite{Seneta2008NonnegativeMA}, the dominant eigenvalue $\lambda_1$ of $P$ is always 1, and its ordered eigenvalues $\lambda_i(P)$ satisfy
\begin{equation}
1=|\lambda_1(P)| \geq |\lambda_2(P)| \geq ... \geq |\lambda_n(P)|.
\end{equation}
Importantly, when the absolute value of the subdominant eigenvalue $\lambda_2(P)<1$, the  Markov chain converges; further, it converges faster with smaller $|\lambda_2(P)|$ \cite{Kirkland2009SubdominantEF}.

It turns out that the subdominant eigenvalues for the matrices in Figure~\ref{fig:markov} are both less than one. However, without blankets, $\lambda_2=0.75$, and with blankets $\lambda_2=0.71$, suggesting that the blanket method should converge faster. Figures~\ref{fig:markov}(c,d) show the actual convergence of the transitions in these two cases, confirming the theory: While without blankets, the process takes 26 steps to converge, only 21 are required with blankets. This result is robust: While the precise number of steps may vary e.g.\ depending on the level of precision in the calculation, the exponential term $0.71^n$ approaches zero more rapidly compared to $0.75^n$ as the value of $n$ increases. As a result, regardless of the specific values or precision used, the blanket method converges more quickly due to this inherent difference in convergence rates dictated by the subdominant eigenvalues of the transition matrices.
Thus the formalization in terms of Markov chains leads to a powerful conclusion in 2-bit AllOnes. The experimental results in Section~\ref{sc:experimental} further suggest that the same conclusion should apply to larger $N$ and to other problems. A challenge for the future is to extend the theory to the general case, as outlined in Section~\ref{sc:discussion}.

\subsection{Distributed Evolution}\label{sc:distribution}

As described in the Background section, although a variety of methods
exist for distributed computing, the hub and spoke model is a
particularly effective approach for evolutionary computation. Through
distribution, it is possible to take advantage of modern computing
resources, including multiple cores on a single machine, several
machines in a LAN (i.e., Local Area Network), or machines in the
cloud. In this section, this method of distribution is first
instantiated in the $(1+1)EA$ algorithm. The approach is then formalized
and an upper bound is derived for the speedup it offers compared to the
non-distributed version.

\subsubsection{Method Description}

With $(1+1)EA$, the hub is a node that stores only a global variable
that maintains the best candidate solution so far. The spokes, or
clients, are nodes running $(1+1)EA$. They regularly communicate with the
hub to possibly obtain a better candidate, or to inform the hub that
they have found such a candidate.  Algorithm~\ref{alg:dist} specifies
these exchanges in detail.

\begin{algorithm}[ht]
   \caption{Hub-and-Spoke distribution (without blankets)}
   \label{alg:dist}
   \begin{algorithmic}
   \STATE {\bfseries Initialization:}
   \bindent
   \STATE Sample $x \in \{0,1\}^N$ uniformly at random and evaluate $f(x)$
   \eindent
   \STATE {\bfseries Optimization:}
   \bindent
   \FOR{$t=1,2,3,...$}
   \STATE {\bfseries Hub interaction:} 
   \bindent
   \STATE Get the hub's candidate $z$
   \IF{$f(x) \geq f(z)$} 
   \STATE Put $x$ as the new hub's candidate
   \ELSE \STATE $x \gets z$
   \ENDIF
   \eindent
   \STATE {\bfseries Offspring generation:} 
   \bindent
   \STATE Calculate mutation rate $\mu$
   \STATE $x^* \gets$ flip each bit of $x$ independently with probability $\mu$
   \eindent
   \STATE {\bfseries Selection:} 
   \bindent
   \IF{$f(x^*) \geq f(x)$} 
   \STATE $x \gets x^*$ 
   \ENDIF
   \eindent
   \ENDFOR
   \eindent
   \end{algorithmic}
\end{algorithm}

\subsubsection{Theoretical Formulation}

Fitness-level theory, also called the fitness-based partitions method,
is widely used for analyzing the runtime of evolutionary algorithms
\cite{Droste2002OnTA}. In this subsection, it is adapted to determining
the upper bound of the computational effort required by the
distribution model compared to a non-distributed evolution.

To begin, the search space is divided into sets $A_1, ..., A_m$,
ordered based on their fitness values. Each set has a lower bound for
the probability of improvement $s_i$, i.e.\ the chance that the search
advances to the next set, i.e.\ to a higher fitness level. Note
that there is no $s_m$ because the global optimum $A_m$ has no room
for improvement.

In elitist evolutionary algorithms such as $(1+1)EA$ (where the
individual with the highest fitness value is always selected for
survival), the best fitness value in the population can only increase.
The set $s_1, ..., s_{m-1}$ can thus be used to calculate an upper
bound on the running time $T$ of the algorithm.  In the case of $(1+1)EA$ it
is equal to the expected number of fitness evaluations:

\begin{equation}
\label{eq:flm}
\begin{aligned}
T \leq \sum^{m-1}_{i=1}\frac{1}{s_i}.
\end{aligned}
\end{equation}

Algorithm~\ref{alg:dist} specifies that all clients are at the same
fitness level almost always (i.e.\ within two hub interactions). Each
client might jump from fitness level $A_i$ to a higher level with
probability $s_i$, i.e.\  each client fails to find an
improvement with a probability of $(1-s_i)$. Because clients are
independent, the probability that all of them fail is
$(1-s_i)^c$. Thus, the probability of leaving $A_i$ is
$d_i=1-(1-s_i)^c$. The upper bound for the running
time of Algorithm \ref{alg:dist} with $c$ clients is then
\begin{equation}
\label{eq:dist}
T \leq \sum^{m-1}_{i=1}\frac{1}{1-(1-s_i)^c}.
\end{equation}

Note that for any $0 \leq x \leq 1$, and any $n > 0$,
\begin{equation}
\label{eq:lemma}
(1-x)^n \leq \frac{1}{1+nx}.
\end{equation}
This inequality \cite{Rowe2012TheCO} can be used to simplify Equation~\ref{eq:dist} and
thus get a better sense of the expected amount of speedup:
\begin{equation*}
\label{eq:dist2}
T \leq \sum^{m-1}_{i=1} \Big[1+\frac{1}{c.s_i}\Big],
\end{equation*}
or simply
\begin{equation*}
\label{eq:dist3}
T \leq (m-1) + \frac{1}{c}\sum^{m-1}_{i=1} \frac{1}{s_i}.
\end{equation*}

This result means that the speedup is linear with more clients. The $(m-1)$ offset becomes negligible for harder problems that have smaller $s_i$ or a smaller number of fitness levels. Note, however, that this $T$ is an upper bound. It is therefore possible that in special cases, the speedup can be higher than the number of clients,
as will be seen later.

\subsection{BLADE}\label{blade-as-a-whole}

The blanket and distribution methods can be combined seamlessly into full BLADE, as described in Algorithm \ref{alg:blade}. This combination is straightforward to implement, and also leads to a surprising synergy, as seen in Section~\ref{sc:synergy}. 

Although BLADE in this paper is implemented for $(1+1)EA$ on binary strings, it can be applied to other population-based evolutionary methods and other representations. Some opportunities are outlined further in Section~\ref{sc:discussion}.

\begin{algorithm}[ht]
   \caption{BLADE (blankets \& hub-and-spoke distribution)}
   \label{alg:blade}
   \begin{algorithmic}
   \STATE {\bfseries Initialization:}
   \bindent
   \STATE Sample $x \in \{0,1\}^N$ uniformly at random and evaluate $f(x)$
   \eindent
   \STATE {\bfseries Optimization:}
   \bindent
   \FOR{$t=1,2,3,...$}
   \STATE {\bfseries Hub interaction:} 
   \bindent
   \STATE Get the hub's candidate $z$
   \IF{$f(x) \geq f(z)$} 
   \STATE Put $x$ as the new hub's candidate
   \ELSE \STATE $x \gets z$
   \ENDIF
   \eindent
   \STATE {\bfseries Offspring generation with blanket:} 
   \bindent
   \STATE Calculate mutation rate $\mu$
   \STATE Sample \emph{blanketLength} from $[1, N-1]$ randomly
   \STATE Set \emph{blanket} $\gets$ flip \emph{blanketLength} random bits in $\{0\}^N$
   \STATE Set $\mu_\mathrm{b} \gets \min(1, \mu(\frac{N}{N-\emph{blanketLength}}))$
   \STATE Sample $y \in \{0,1\}^N$ at random with $p(1)=\mu_\mathrm{b}$
   \STATE Set $\emph{blanket} \gets \emph{blanket} \land y$ ($\land$ is bitwise AND)
   \STATE Create $x^* \gets x \oplus$ \emph{blanket} ($\oplus$ is bitwise XOR)
   \eindent
   \STATE {\bfseries Selection:} 
   \bindent
   \IF{$f(x^*) \geq f(x)$} 
   \STATE $x \gets x^*$ 
   \ENDIF
   \eindent
   \ENDFOR
   \eindent
   \end{algorithmic}
\end{algorithm}

\section{Experimental Analysis}
\label{sc:experimental}

BLADE was evaluated in three benchmark problems, selected to cover domains with a variety of different fitness landscapes. Experiments were performed on each to evaluate the contribution of masking and distribution separately and then together.

\subsection{Experimental Setup}\label{allones}

The first problem, AllOnes, is an optimization problem where the goal is to find a binary string of length $N$ that consists of all ones. The fitness landscape is a needle in a haystack: While it is easy to identify the global optimum, it is hard to find it among all the solutions. There is only one combination of the $N$ bits that has the fitness of one and all the remaining $(2^N - 1)$ combinations have the fitness of zero. Solving AllOnes is analogous to searching through the $2^N$ possibilities to open a binary combination lock.

The second problem, OneMax (or Hamming distance)\cite{Doerr_onemax:gecco2019}, is a classic problem in evolutionary computation, and is widely used to evaluate the performance of optimization algorithms. In OneMax, the goal is also to find a binary string where all the bits are one, but the fitness of a candidate solution is equal to the number of ones in the string. OneMax is a relatively easy problem for evolutionary computation, however, it is a useful baseline often referred to as the “drosophila of evolutionary computation” \cite{Doerr_onemax:gecco2020}

The third problem, LeadingOnes, is another classical problem widely used to evaluate the performance of optimization algorithms. The goal is, again, to find a binary string where all the bits are one, but in this case, the fitness of a candidate solution is equal to the number of ones at the beginning of the string. An advantage of using this problem for benchmarking evolutionary computation methods is that its theoretical convergence bounds for both static and adaptive mutation rates are known for $(1+1)EA$ \cite{doerr-papa:gecco2020}. Therefore, it is possible to compare empirical results against them to establish the baseline.

Each optimization problem was run a thousand times; the reported convergence numbers are the averages for those runs. Further, 95\% confidence bounds were calculated to measure the statistical significance of the results. In AllOnes, convergence time increases exponentially with the size of the problem, and therefore the string lengths of 2 to 16 bits were used. In OneMax and LeadingOnes, experiments were run from 2 to 32 bits.

The same mutation rates were used as a basis for all runs; BLADE
modifies it on the fly based on the length of its random blanket
(according to Algorithms~\ref{alg:blanket} and~\ref{alg:blade}). In
AllOnes and OneMax, the static rate of $\frac{1}{N}$ was used. In
LeadingOnes, there were two cases: $\frac{1.5936}{N}$ was used in the
static case, and $\frac{1}{1+\mathrm{LO}(x)}$, where $\mathrm{LO}(x)$ is the fitness of
the candidate $x$, in the adaptive case. These are the theoretical
optimum rates for this domain \cite{doerr-papa:gecco2020,Doerr_onemax:gecco2020}.

The results for the masking component alone, i.e.\ BLADE on a single
client, are described in the next subsection, followed by experiments
on evaluating the effect of distributing BLADE over two to eight
clients. The final subsection analyzes the speedups resulting from
distribution, identifying a surprising synergy between blankets and
distribution 

\subsection{Experiments With Blankets}\label{blanket-experiments}

\begin{figure*}[t]
  \centering
  \includegraphics[width=0.48\linewidth]{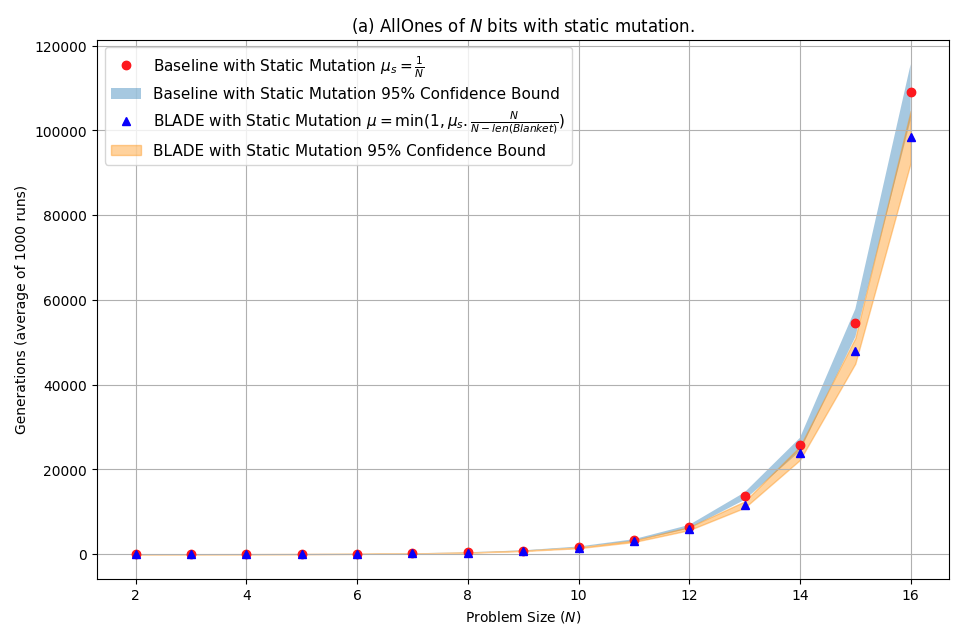}
  \includegraphics[width=0.48\linewidth]{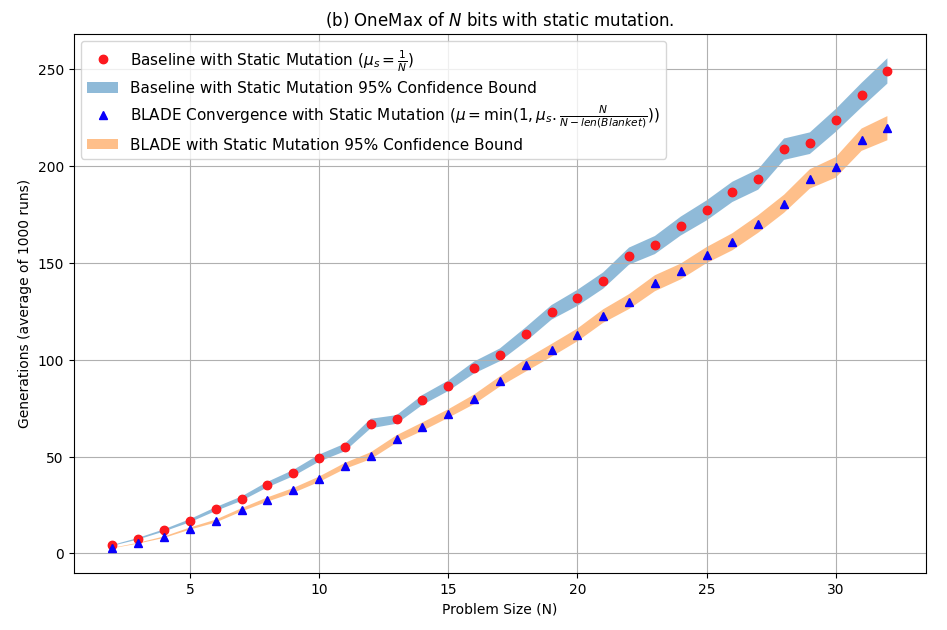}
  \includegraphics[width=0.48\linewidth]{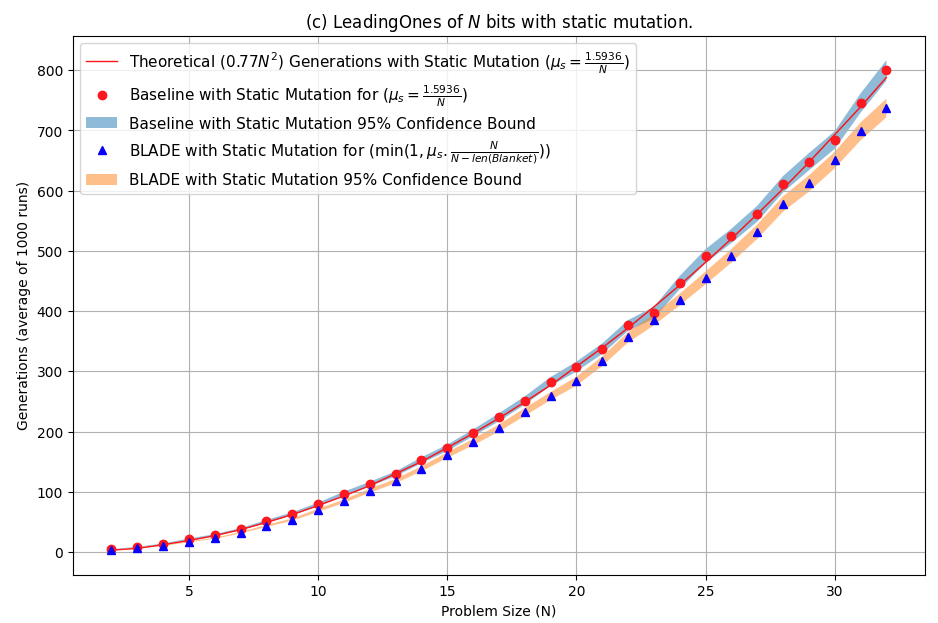}
  \includegraphics[width=0.48\linewidth]{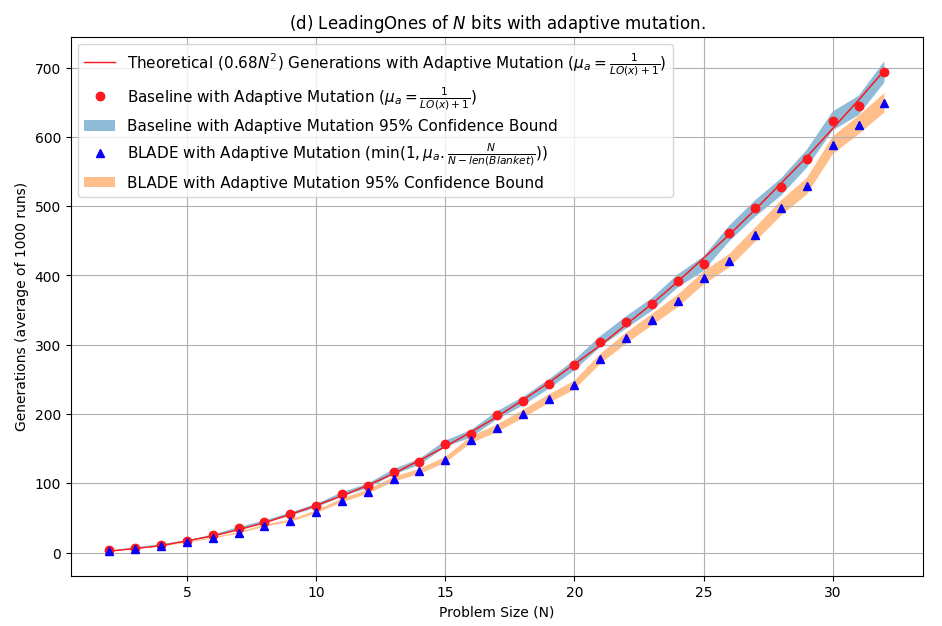}
  \caption{A comparison between the blanket method (i.e.\ BLADE on a single host) and the baseline method (i.e.\ standard evolution without blankets) across four benchmark problems: (a) AllOnes, (b) OneMax, and LeadingOnes with (c) static mutation and (d) adaptive mutation. The mutation rate for AllOnes and OneMax was $\mu_=\frac{1}{N}$. For LeadingOnes, the theoretically optimal mutation rates were used, i.e.\, $\mu=\frac{1.5936}{N}$ in the static case and $\mu=\frac{1}{\mathrm{LO}(x)+1}$ in the adaptive case. The blanket method modifies these mutation rates by a factor of $\frac{N}{N-\mathrm{len(blanket)}}$ and clips them to one. The $x$-axis denotes the problem size $N$ (i.e.\ the length of the binary string) and the $y$-axis the average number of generations to converge, averaged over 1000 runs. The shaded areas indicate 95\% confidence intervals. The results show that blankets improve convergence significantly on all problems.}
  \label{fig:masking}
\end{figure*}

The blanket technique was first implemented and evaluated on a single client, without distribution. A summary of the results is shown in Figure \ref{fig:masking}; a detailed discussion follows for each benchmark problem.

\subsubsection{AllOnes}
Figure~\ref{fig:masking}(a) illustrates the advantage of using blankets on
AllOnes.  Given that this is a needle-in-a-haystack problem, and the
search space grows exponentially with string length, it is no surprise
that the convergence time increases exponentially as well. However,
BLADE converges slightly faster than the baseline, presumably due to
the subdominant eigenvalues of the corresponding transition matrices.
Further, as the problem size increase, the advantage of blankets
becomes more pronounced.

\subsubsection{OneMax}\label{onemax-single}
Figure~\ref{fig:masking}(b) shows the advantage of using blankets on
OneMax.  BLADE converges significantly faster than the
baseline, as indicated by a wide separation of the 95\% confidence
bounds, and the difference increases with problem size.

\subsubsection{LeadingOnes}
Figure~\ref{fig:masking}(c) compares BLADE with the static mutation
baseline and Figure~\ref{fig:masking}(d) with the adaptive
mutation baseline on LeadingOnes. Again, BLADE converges significantly
faster than the baseline in both cases, and the advantage increases
with problem size.

With the theoretically optimal mutation rate, the convergence rate
with static mutation is $0.77N^2$, and with adaptive mutation is
$0.68N^2$ \cite{doerr-papa:gecco2020,Doerr_onemax:gecco2020}.  These
rates are plotted as continuous lines in Figures
\ref{fig:masking}(c,d). As expected, they match the experimental
results well.

\subsection{Experiments With Distribution}

\begin{figure*}[t]
  \centering
  \includegraphics[width=0.48\linewidth]{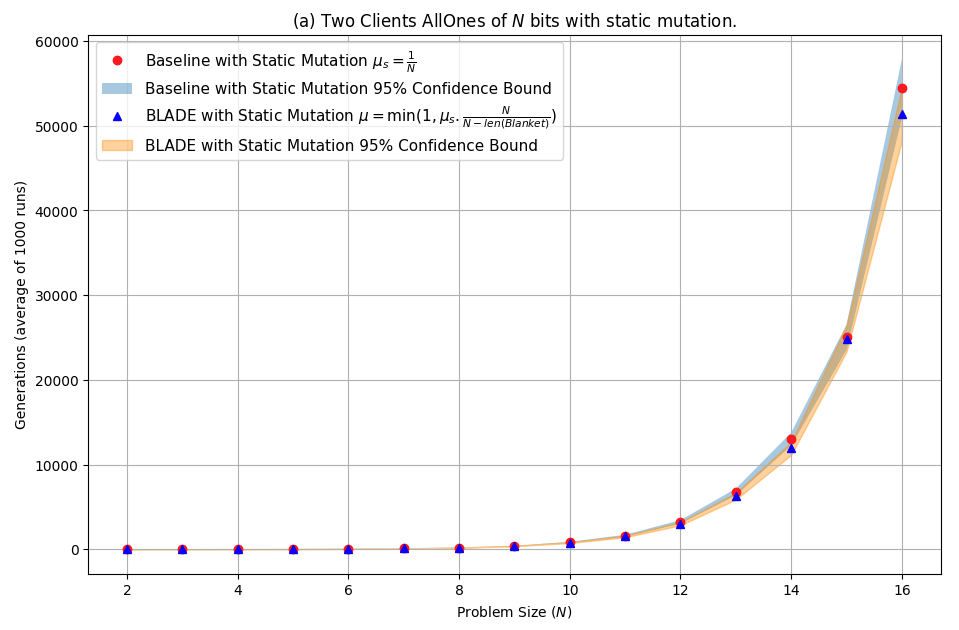}
  \includegraphics[width=0.48\linewidth]{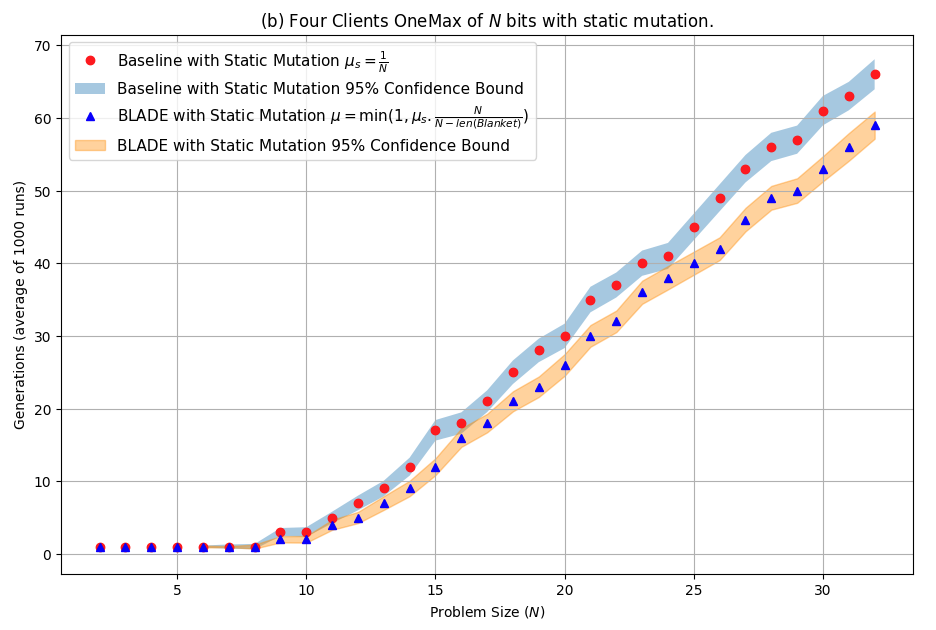}
  \includegraphics[width=0.48\linewidth]{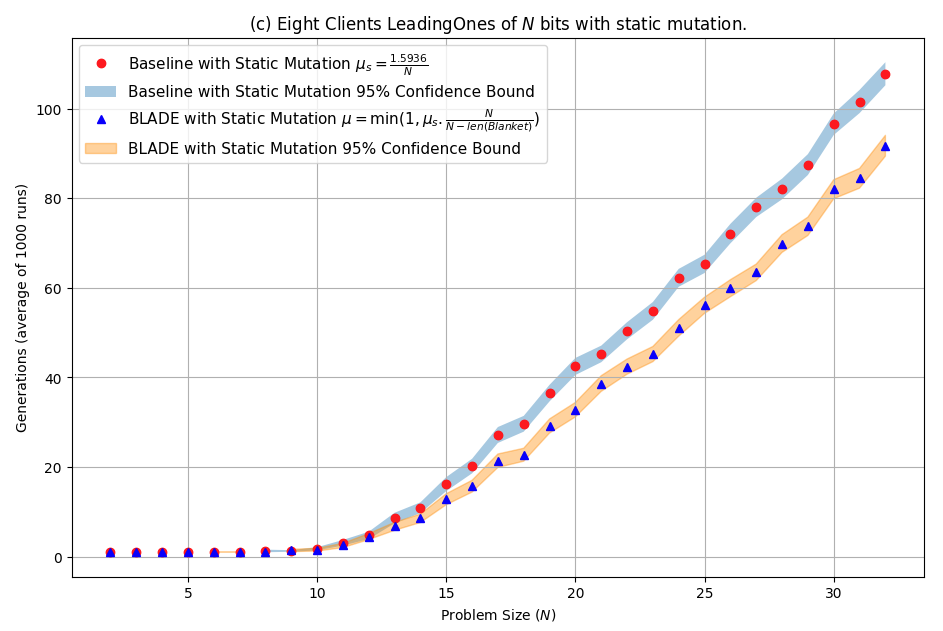}
  \includegraphics[width=0.48\linewidth]{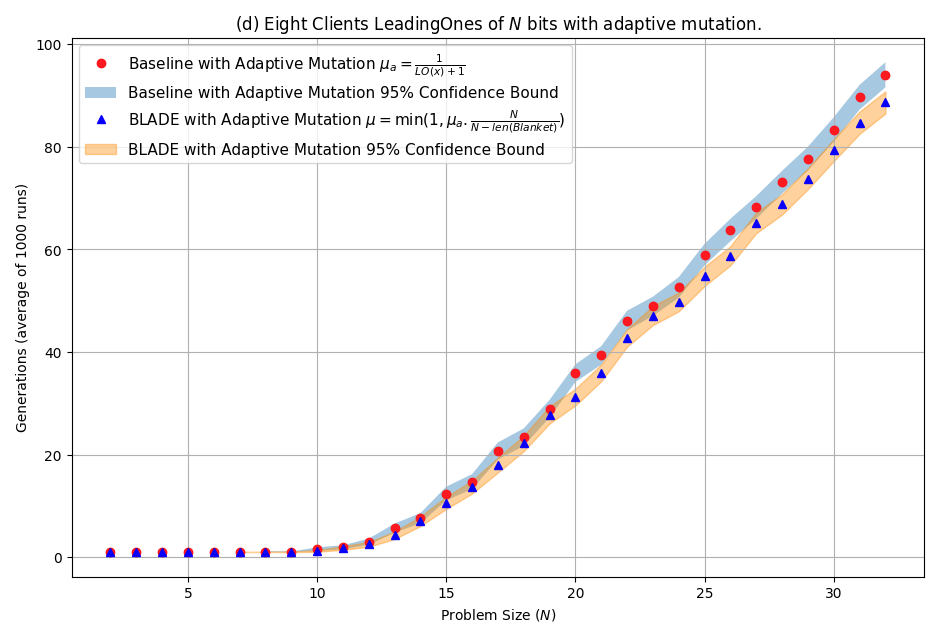}
  \caption{A comparison between full BLADE (including both blankets and distribution methods) and the baseline across the four benchmark problems. Representative results with two, four, and eight clients are shown; the complete set is included in Appendix~\ref{app1}. The experimental and display details are the same as in Figure~\ref{fig:masking}. The advantage of blankets extends to distribution across several clients: BLADE converges significantly faster than the baseline in all cases.}
  \label{fig:distribution}
\end{figure*}

The aim of these experiments is to study the contrast between just distributing the problem set and utilizing both blanket and distribution as BLADE does. Experiments were conducted for two, four, and eight clients and the results are depicted in Figure \ref{fig:distribution}. A thorough discussion for each problem set is provided. In addition, a complete collection of these graphs can be found in \ref{app1} for further examination.

\subsubsection{AllOnes}
Figure~\ref{fig:distribution}(a) shows the advantage of using BLADE
(combining distribution and masking) on AllOnes. The results are
similar to the single client case: Both are exponential and BLADE is
slightly better, with an increasing difference.  Similar results were
obtained in the four and eight-client cases.

\subsubsection{OneMax}
Figure~\ref{fig:distribution}(b) shows the advantage of BLADE on
OneMax with distribution over four clients.  Again, the results are
similar to the single-client case, with BLADE converging
significantly and increasingly faster than the baseline.  Similar results were
obtained in the two and eight-client cases.

\subsubsection{LeadingOnes}
Figures~\ref{fig:distribution}(c,d) compares BLADE with baseline on
LeadingOnes distributed over eight clients.  Again, BLADE converges
significantly and increasingly faster than the baseline.  Similar results were obtained in the two and four-client cases.

%One might still argue that the mutation rate modification by BLADE will put it in the category of adaptive mutation although the Blanket is random and the modification factor of $\frac{N}{N-len(Blanket}$ is domain-independent. A good answer to that argument is to apply BLADE on top of the theoretical optimum adaptive mutation rate of $\frac{1}{LO(x)+1}$ for LeadingOnes and see whether we get an improvement there too. This is exactly what is shown in Figure \ref{fig:distribution}(d)

\subsection{Synergy of Blankets and Distribution}
\label{sc:synergy}

Previous sections demonstrated that using blankets improves
search performance over baseline both when it is run on a single client and when it is distributed over several clients. An interesting question is: Is there a synergy between blankets and distribution? That is, does distribution offer a larger speedup with BLADE than it does with the baseline?

\begin{figure*}[t]
  \centering
  \includegraphics[width=0.48\linewidth]{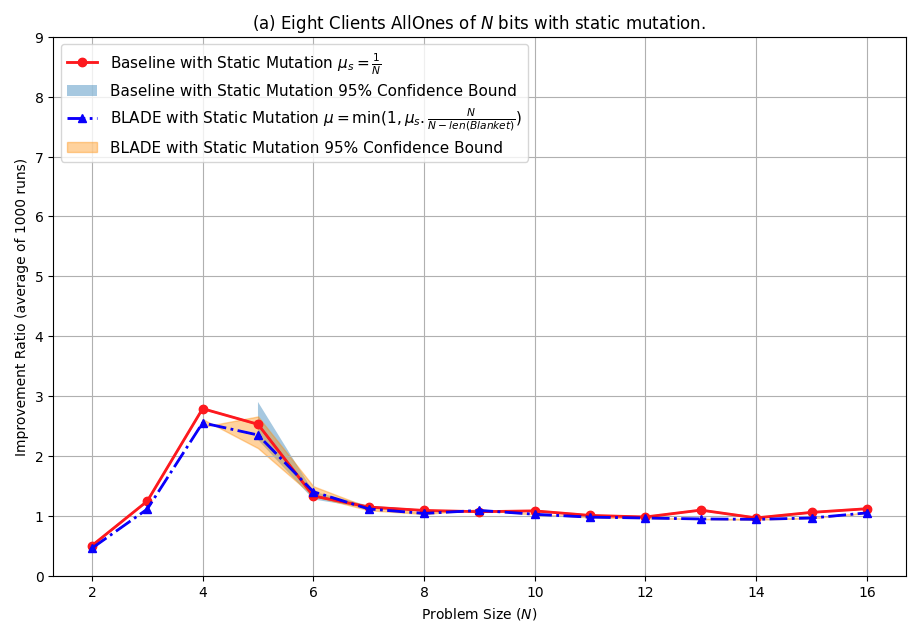}
  \includegraphics[width=0.48\linewidth]{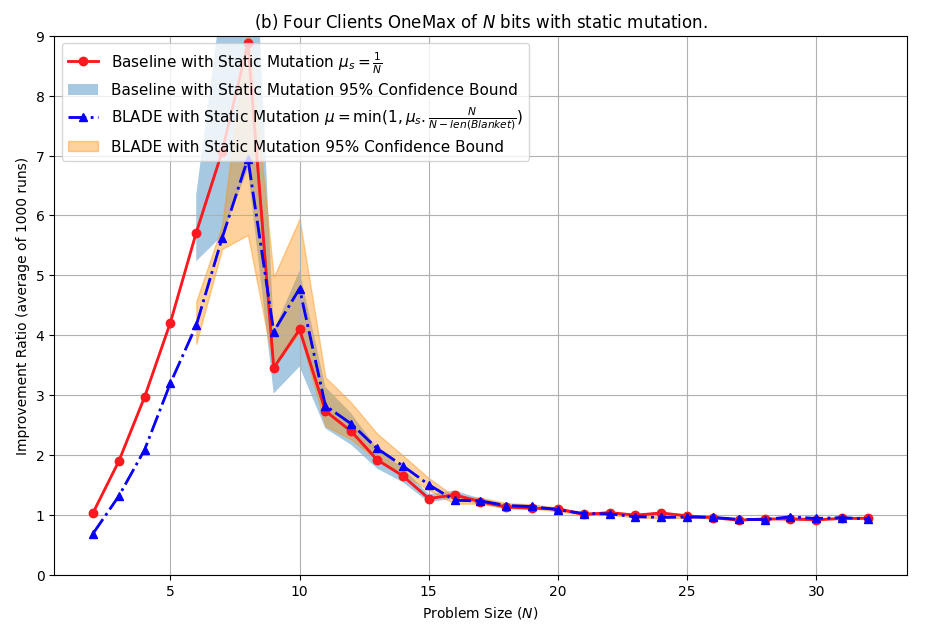}
  \includegraphics[width=0.48\linewidth]{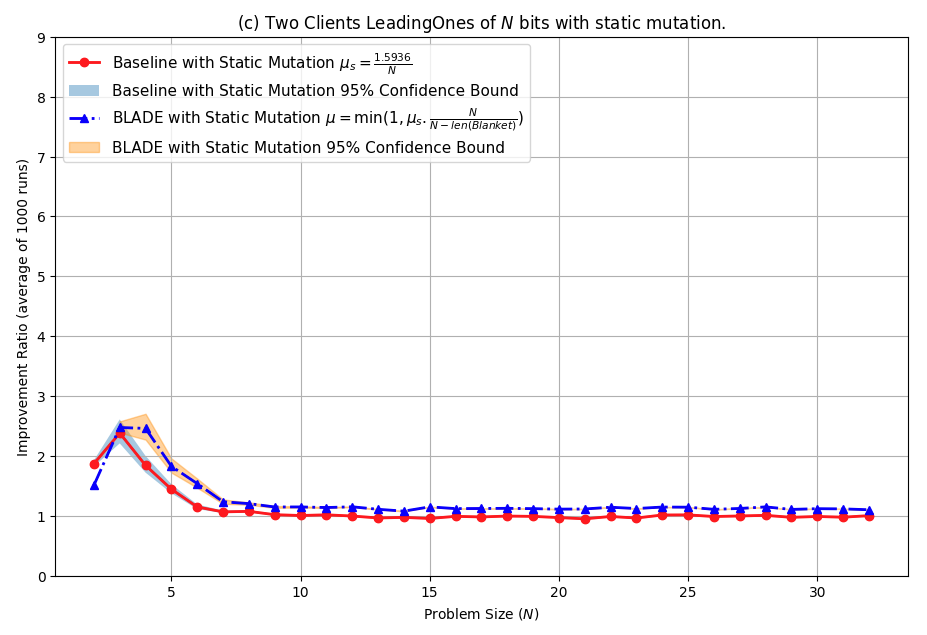}
  \includegraphics[width=0.48\linewidth]{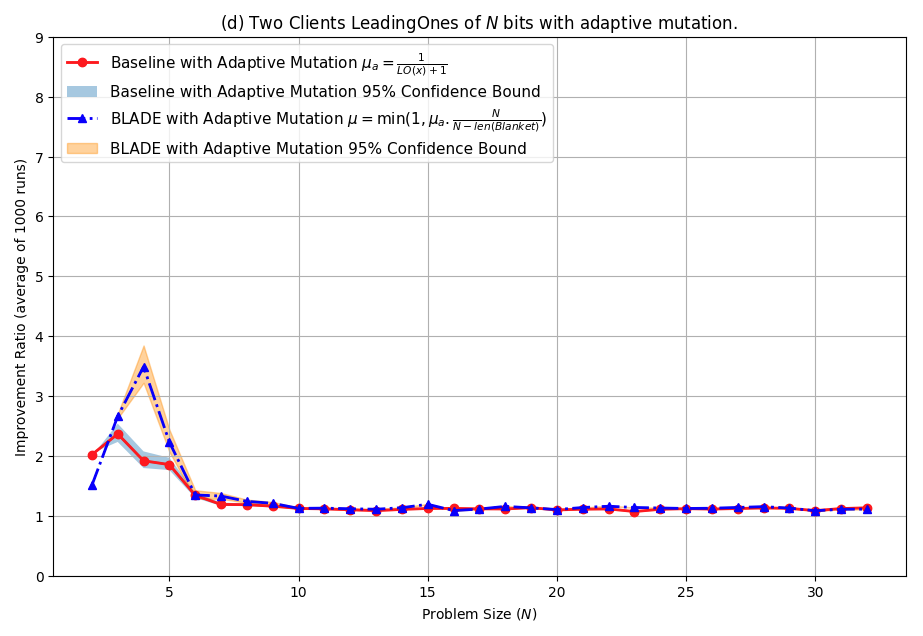}
  \caption{A comparison of the speedup ratio of BLADE vs.\ the baseline on the four benchmark problems. Similarly to Figure~\ref{fig:distribution}, representative results with two, four, and eight clients are shown; the complete set is included in Appendix~\ref{app2}.  A ratio of 1.0 indicates that the distribution is perfectly efficient, i.e.\ the total number of evaluations across all clients is the same as the number of evaluations on a single client. As the problem size grows, the ratio approaches 1.0 for AllOnes and OneMax. Remarkably, for LeadingOnes the ratio is above 1.0 for both the baseline and BLADE with adaptive mutation (d), and for BLADE only with static mutation (c). The results thus suggest that there is a synergy between adaptive mutation and distribution, and that BLADE provides the crucial adaptation in the otherwise static mutation case.}
  \label{fig:synergy}
\end{figure*}

To answer this question, the ratios of total evaluations in single-client
and multi-client runs are plotted in Figure \ref{fig:synergy} for
representative cases in each benchmark. A ratio of 1.0 means that the
speedup is perfectly efficient, e.g.\ a run distributed over two
clients converges twice as fast as a run on a single client. A ratio
above one means that the threads provide additional information 
that the distribution algorithm can utilize to speed up the search
even more.

A summary of these results is given below, and the comprehensive set
of plots is included in Appendix~ \ref{app2}.

\vspace*{-1ex}
\subsubsection{AllOnes}

Figure~\ref{fig:synergy}(a) shows the improvement ratios in AllOnes
when the runs are distributed over eight clients. When $N$ is small,
there is a non-negligible chance that some of the clients have
high-fitness individuals in their initial population, resulting in the
high ratios up to $N=8$. With larger $N$, the ratios are close to
one, suggesting that the distribution is efficient, and both the
baseline and BLADE benefit from it equally. Similar results were
obtained in the two and four-client cases.

\subsubsection{OneMax}
Figure~ \ref{fig:synergy}(b) illustrates the improvement ratios in
OneMax with a four-client distribution. The chance of having high-fitness
individuals in the initial population is higher in this problem, and
has a significant effect up to $N=20$. With larger $N$, the ratios are
again close to one for both the baseline and BLADE, as in AllOnes.
Similar results were obtained in the two and eight-client cases.

\subsubsection{LeadingOnes}
Figures~\ref{fig:synergy}(c,d) plots the improvement ratios in
LeadingOnes with static and adaptive mutation with a two-client
distribution. In this benchmark, the effect of lucky initialization is
again small, and negligible after about $N=8$. With larger $N$, an
interesting observation can be made: The improvement is significantly
greater than one for BLADE in the static case, and for both the
baseline and BLADE in the adaptive case. Similar results were obtained
in the four and eight-client cases.

Apparently, in LeadingOnes there is information in the two threads
that can be utilized to improve the search.  This information can be
captured to an extent through adaptive mutation; however, even when
the mutation is static (as in Figure~\ref{fig:synergy}(c)), BLADE can still capture it. BLADE adjusts its
mutation based on the blankets, and therefore establishes a version of
the adaptive mutation process. This process allows it to take
advantage of the synergy between threads more effectively than the
baseline. Characterizing and optimizing this mechanism is a most
exciting direction for future work.

\vspace*{-1ex}
\section{Discussion and Future Work}
\label{sc:discussion}
% \begin{itemize}
%     \item Applying BLADE is not cumbersome while it might entertain the synergy between using blankets and distribution in some problem domains and give us more than n-fold speedups.
%     \item BLADE could be used as a plug-in to the method of choice.
%     \item BLADE can easily expand to population-based evolutionary methods.
%     \item Despite the relevance to binary problems, blankets could be abstracted to any arbitrary construction blocks for different units of evolution (e.g., $N$-dimensional vector optimization).
%     \item It would be nice to apply BLADE to a real-world application and showcase its effectiveness (e.g., Stanford brain modeling project).
%     \item There might be different angles for a mathematical proof.
% \end{itemize}

BLADE can potentially be used to accelerate evolutionary algorithms by utilizing blanket-based tuning of search and by distributing the search. The experiments covered a wide range of fitness landscapes representative of many practical problems. The method is easily integrated as a plug-in into any preferred evolutionary method. 

In order to put these conclusions into practice, there are two immediate directions for future work. The first is to extend the method from a $(1+1)EA$ to a population-based approach; the second is to generalize blankets to other evolutionary representations such as multi-dimensional vectors and trees. Once the BLADE method is extended in this manner, it can be tested in real-world applications. The goal will be to verify that more than $n$-fold speedup can be obtained with $n$ clients, taking advantage of the synergy between its two components.

Future theoretical research may seek to generalize the Markov chain approach to other problems and sizes. A particularly interesting challenge is to identify the conditions under which the synergy can emerge, and derive bounds for it. Such an understanding could be instrumental in developing faster evolutionary computation implementations in the future.

% \newpage
\vspace*{-1ex}
\section{Conclusion}

BLADE was demonstrated to accelerate a fundamental evolutionary algorithm in several benchmark problems. It can be easily integrated into other existing algorithms, making it possible to take advantage of it in practical applications. Its potential for providing more than $n$-fold speedup with $n$ clients is particularly intriguing and worthy of further study.

% \begin{acks}
% To ...
% \end{acks}

\newpage
\bibliographystyle{ACM-Reference-Format}
\bibliography{main}

\newpage
\onecolumn
\appendix

\section{Appendix: Comprehensive sets of graphs}

\subsection{Distribution Graphs}\label{app1}

\begin{figure*}[ht]
  \centering
  \includegraphics[width=0.48\linewidth]{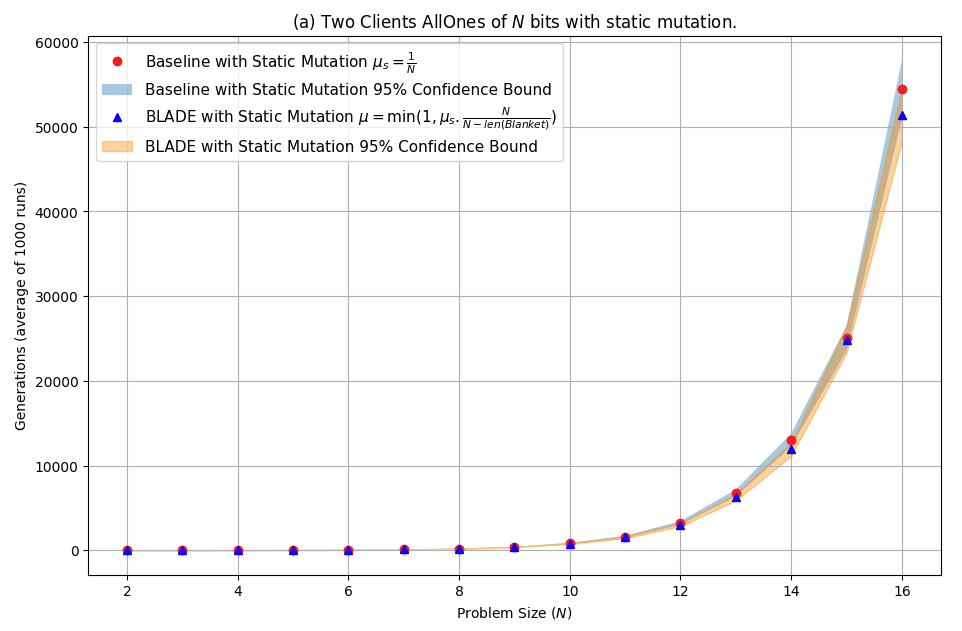}
  \includegraphics[width=0.48\linewidth]{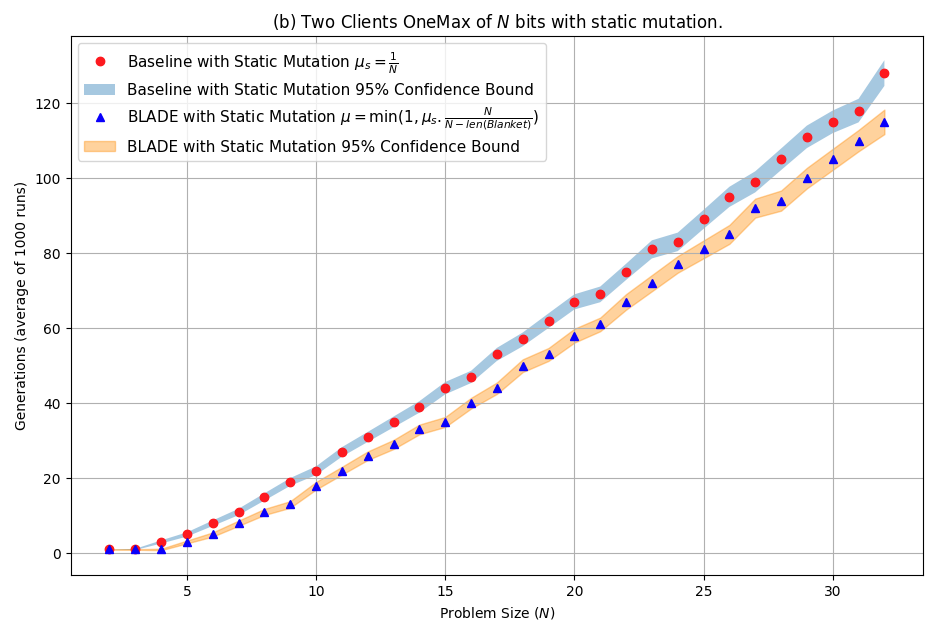}
  \includegraphics[width=0.48\linewidth]{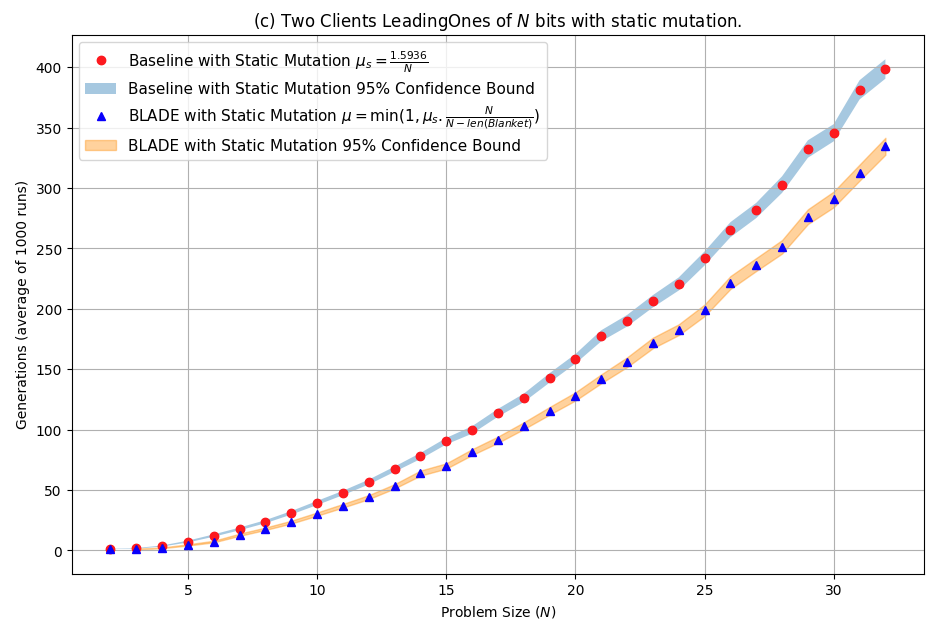}
  \includegraphics[width=0.48\linewidth]{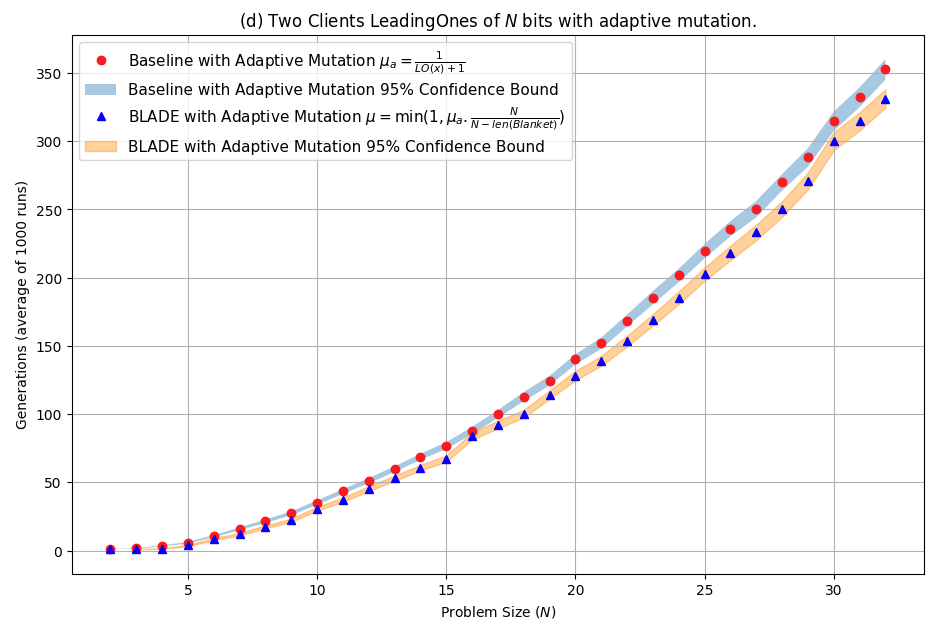}
  \caption{A comparison between BLADE and the baseline across the four benchmark problems with distribution over \emph{two} clients; the experimental and display details are the same as in Figure~\ref{fig:distribution}. BLADE converges significantly faster than the baseline.}
  \Description{}
  \label{fig:A-1-1}
\end{figure*}

\begin{figure*}[ht]
  \centering
  \includegraphics[width=0.48\linewidth]{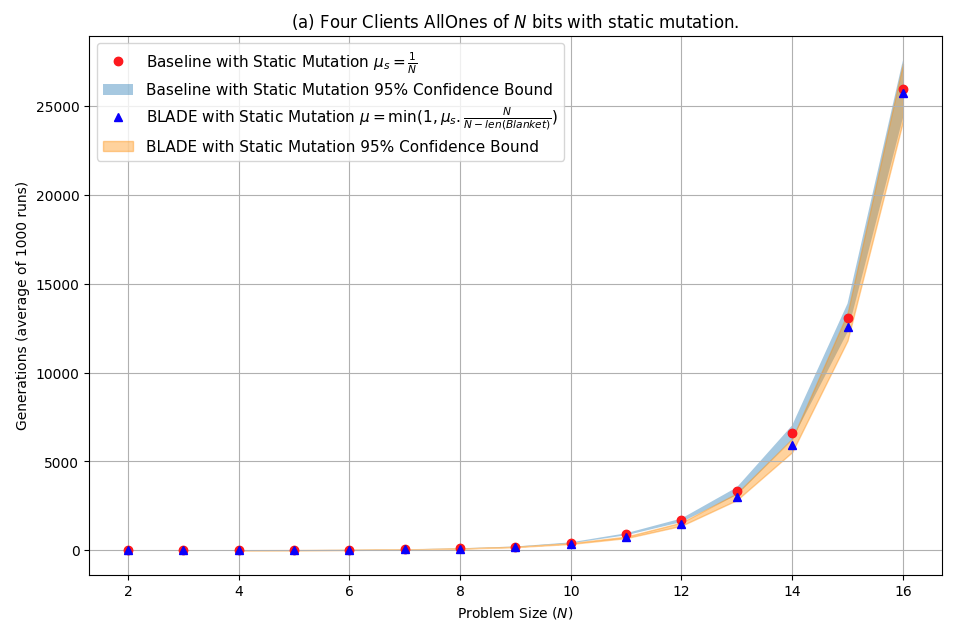}
  \includegraphics[width=0.48\linewidth]{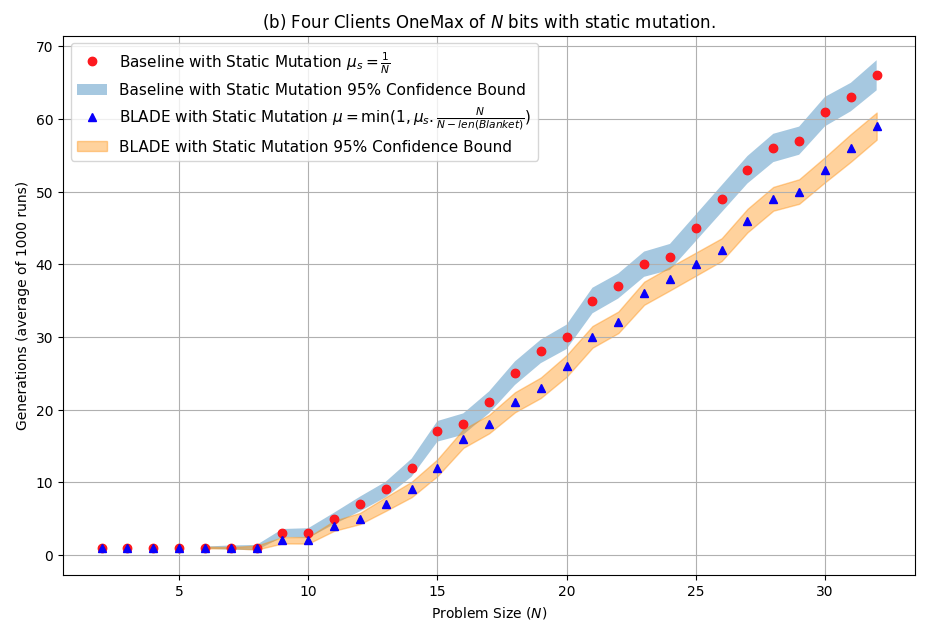}
  \includegraphics[width=0.48\linewidth]{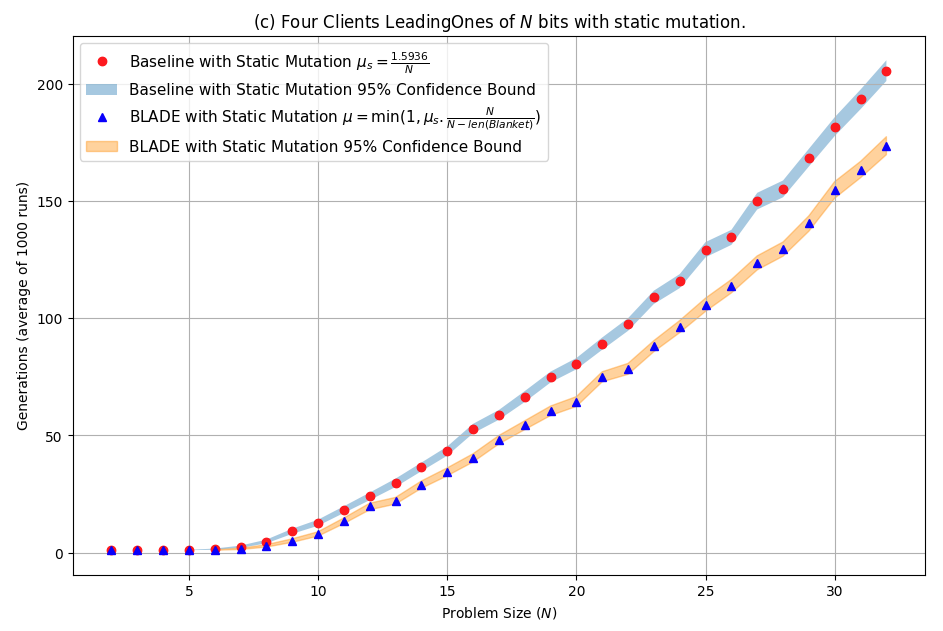}
  \includegraphics[width=0.48\linewidth]{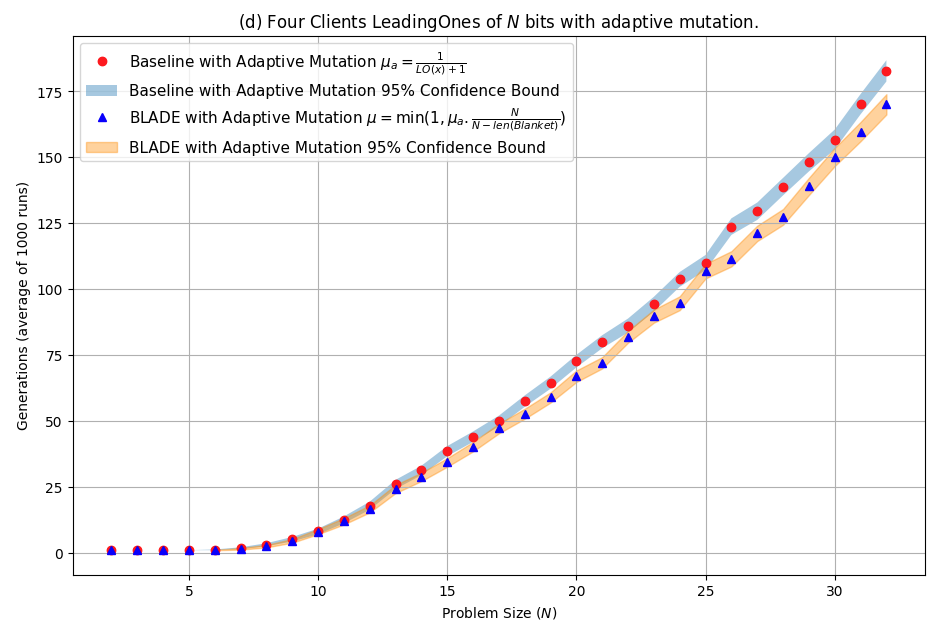}
  \caption{A comparison between BLADE and the baseline across the four benchmark problems with distribution over \emph{four} clients; the experimental and display details are the same as in Figure~\ref{fig:distribution}. BLADE converges significantly faster than the baseline.}
  \Description{}
  \label{fig:A-1-2}
\end{figure*}

\begin{figure*}[ht]
  \centering
  \includegraphics[width=0.48\linewidth]{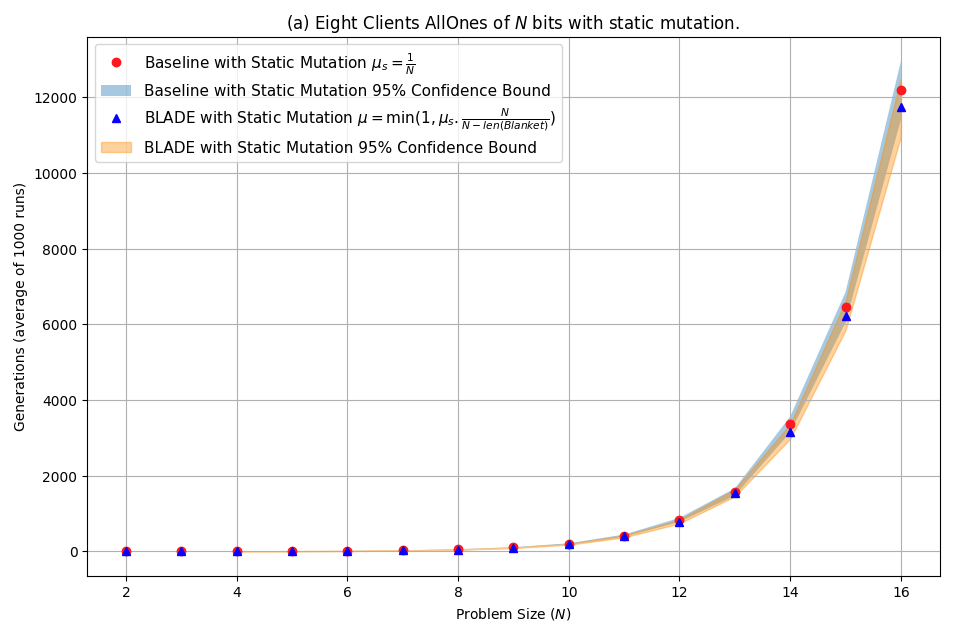}
  \includegraphics[width=0.48\linewidth]{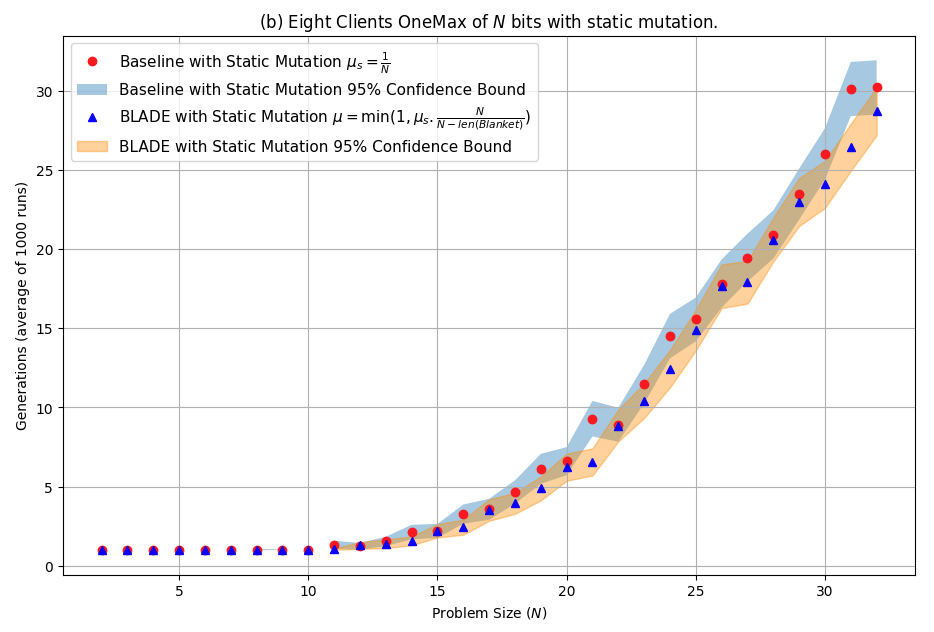}
  \includegraphics[width=0.48\linewidth]{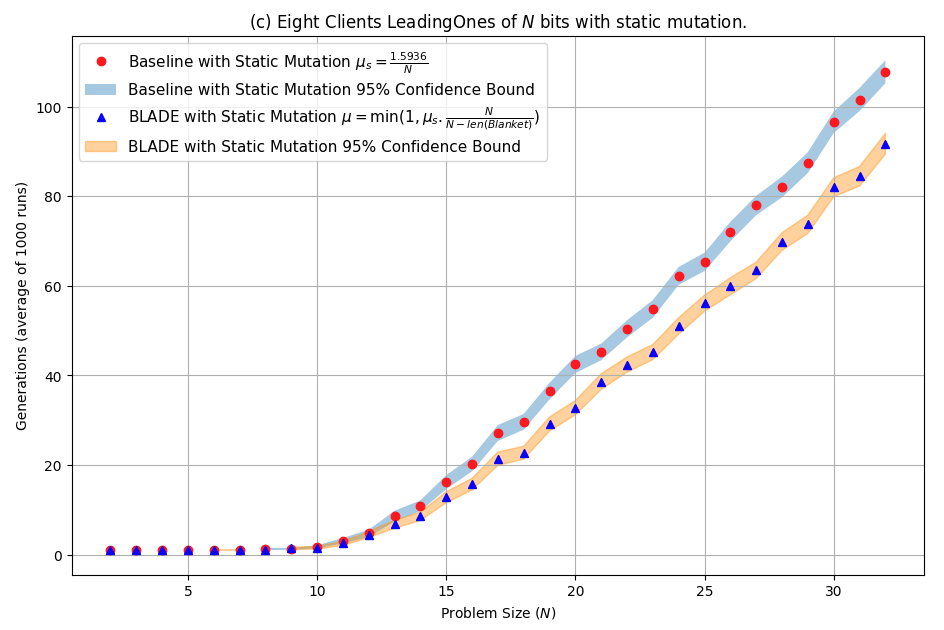}
  \includegraphics[width=0.48\linewidth]{figures/A.1.3-d.png}
  \caption{A comparison between BLADE and the baseline across the four benchmark problems with distribution over \emph{eight} clients; the experimental and display details are the same as in Figure~\ref{fig:distribution}. BLADE converges significantly faster than the baseline.}
  \Description{}
  \label{fig:A-1-3}
\end{figure*}

Figures~\ref{fig:A-1-1}, \ref{fig:A-1-2}, and~\ref{fig:A-1-3} show the results of the comparisons between BLADE and the baseline on the four benchmark problems with two, four, and eight clients, respectively. 

\clearpage
\subsection{Synergy Graphs}\label{app2}

\begin{figure*}[htb]
  \centering
  \includegraphics[width=0.48\linewidth]{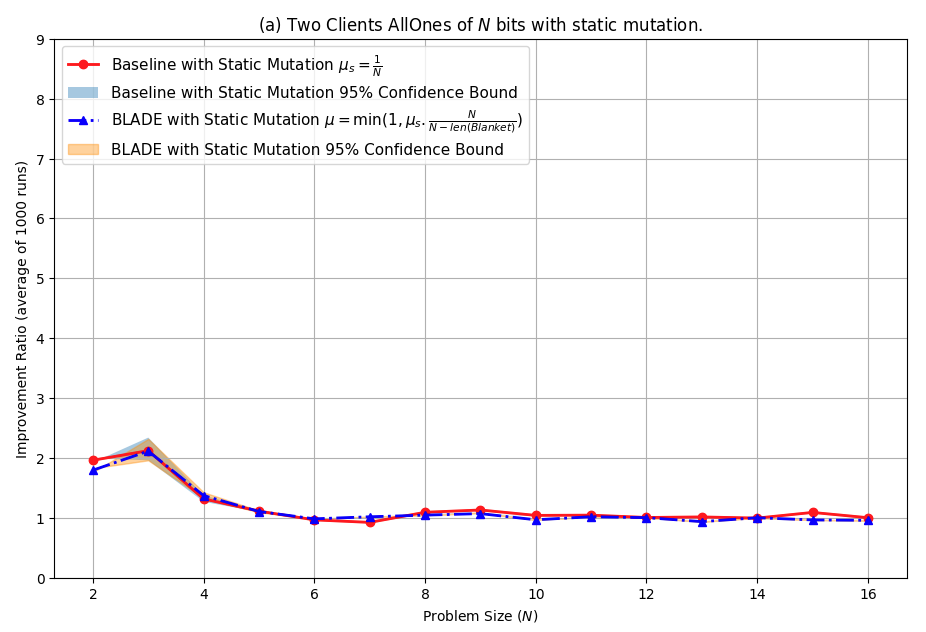}
  \includegraphics[width=0.48\linewidth]{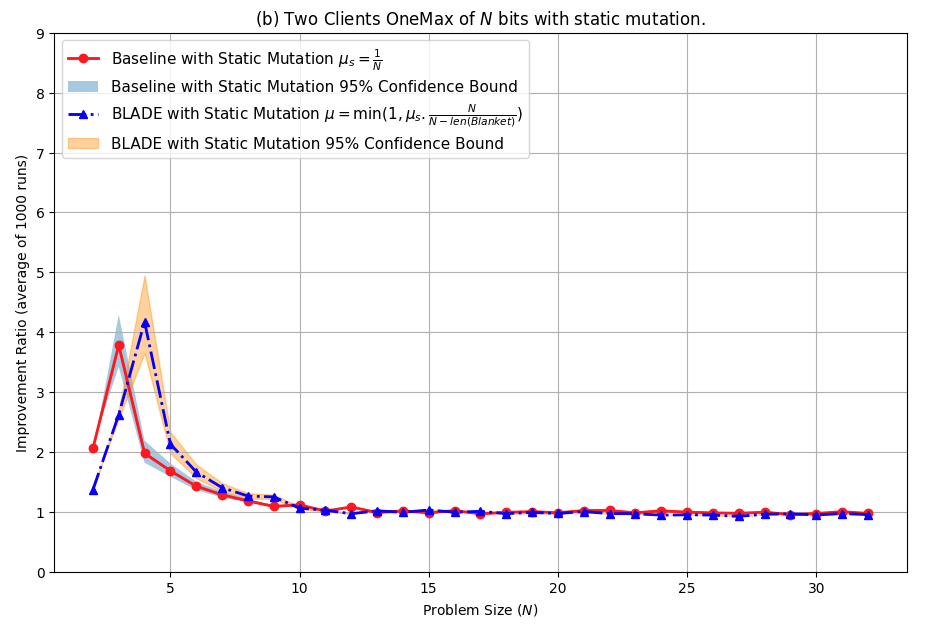}
  \includegraphics[width=0.48\linewidth]{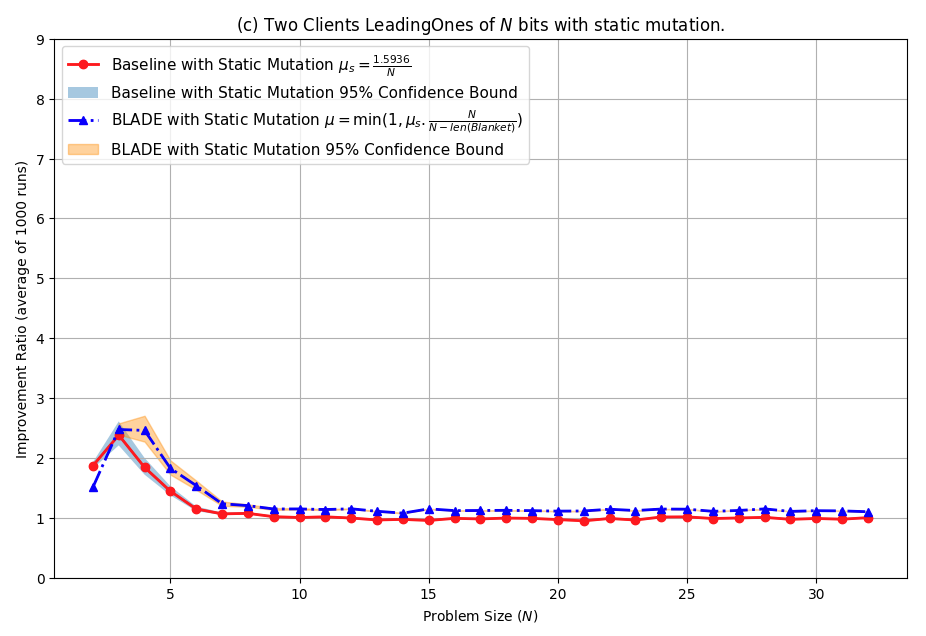}
  \includegraphics[width=0.48\linewidth]{figures/A.2.1-d.png}
  \caption{A comparison of the speedup ratio of BLADE vs.\ the baseline on the four benchmark problems with distribution over \emph{two} clients; the experimental and display details and conclusions are similar to those in Figure~\ref{fig:synergy}.}
  \Description{}
  \label{fig:A-2-1}
\end{figure*}

\begin{figure*}[ht]
  \centering
  \includegraphics[width=0.48\linewidth]{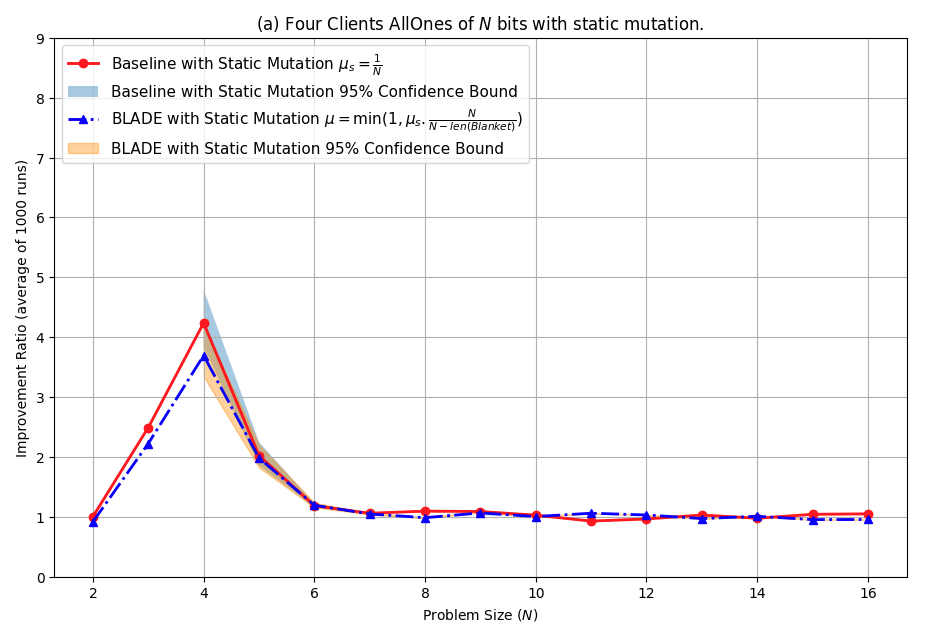}
  \includegraphics[width=0.48\linewidth]{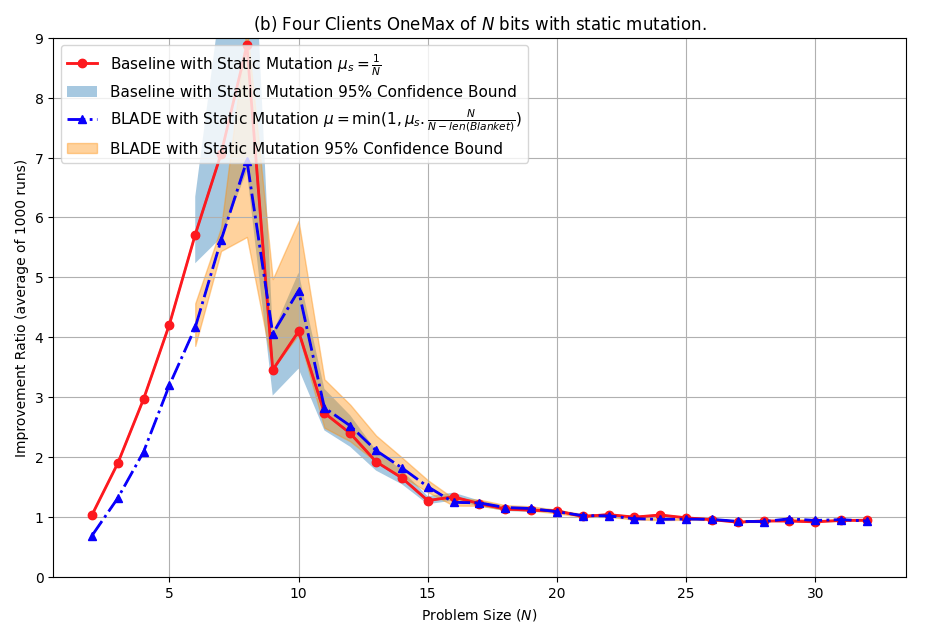}
  \includegraphics[width=0.48\linewidth]{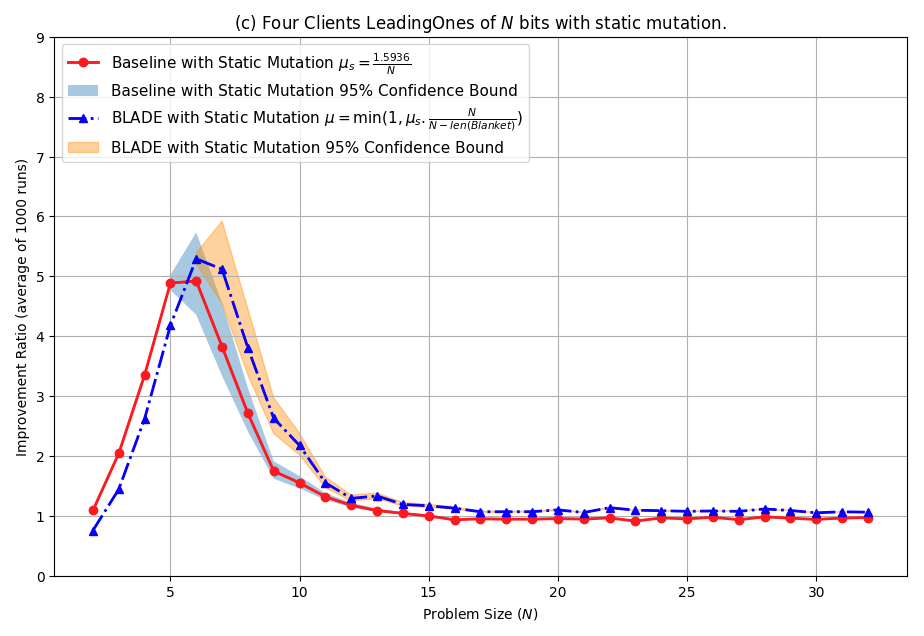}
  \includegraphics[width=0.48\linewidth]{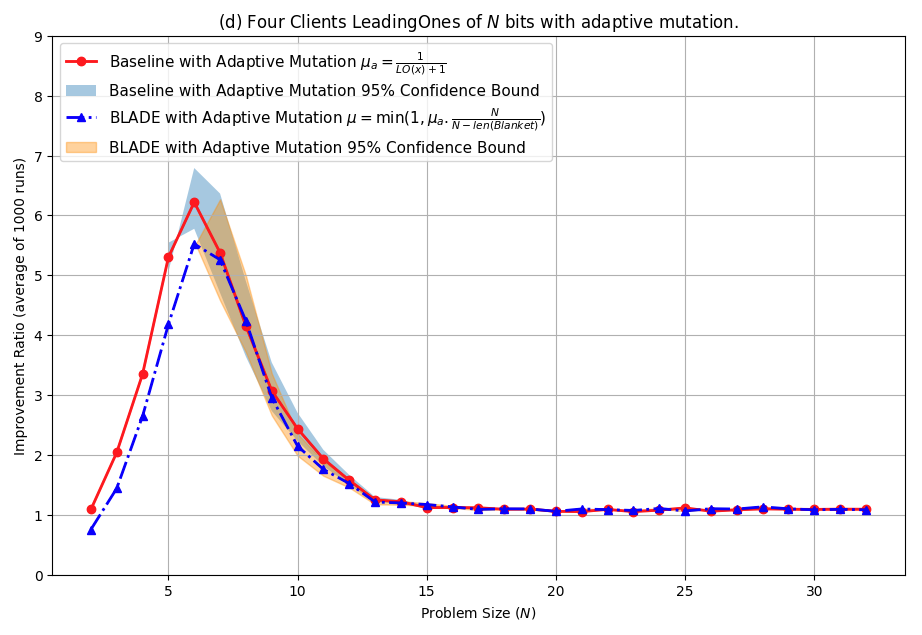}
  \caption{A comparison of the speedup ratio of BLADE vs.\ the baseline on the four benchmark problems with distribution over \emph{four} clients; the experimental and display details and conclusions are similar to those in Figure~\ref{fig:synergy}.}
  \Description{}
  \label{fig:A-2-2}
\end{figure*}

\begin{figure*}[ht]
  \centering
  \includegraphics[width=0.48\linewidth]{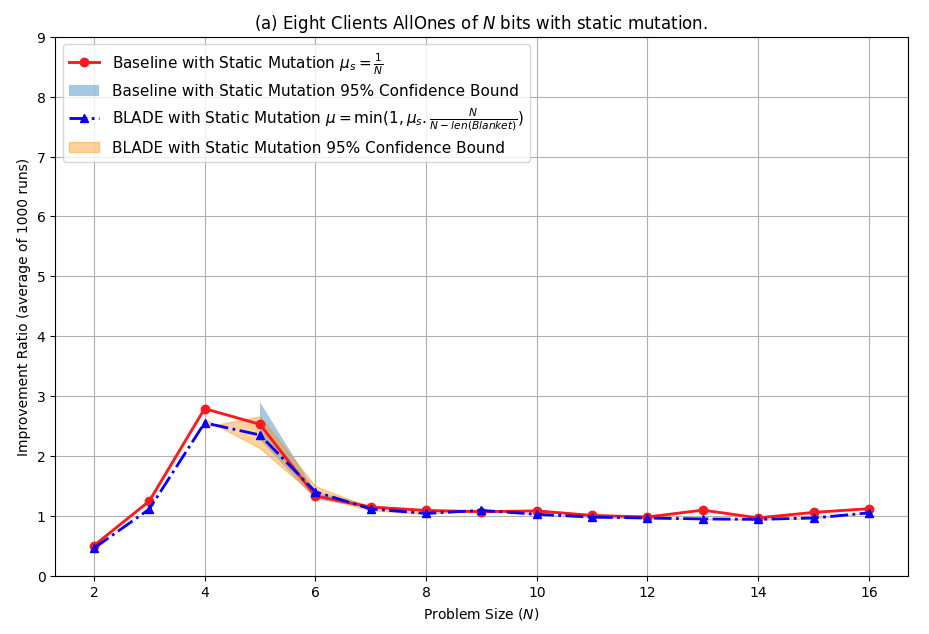}
  \includegraphics[width=0.48\linewidth]{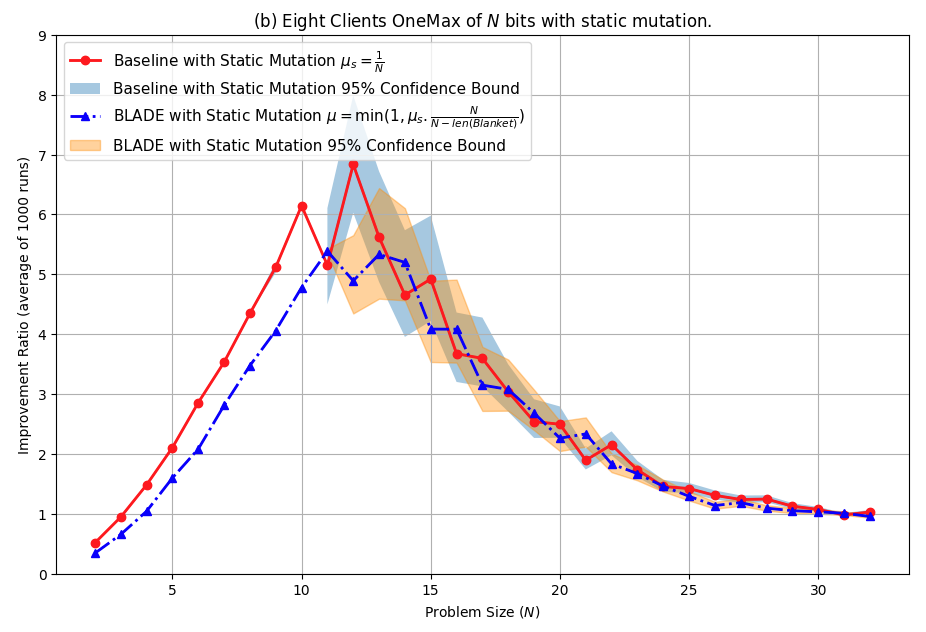}
  \includegraphics[width=0.48\linewidth]{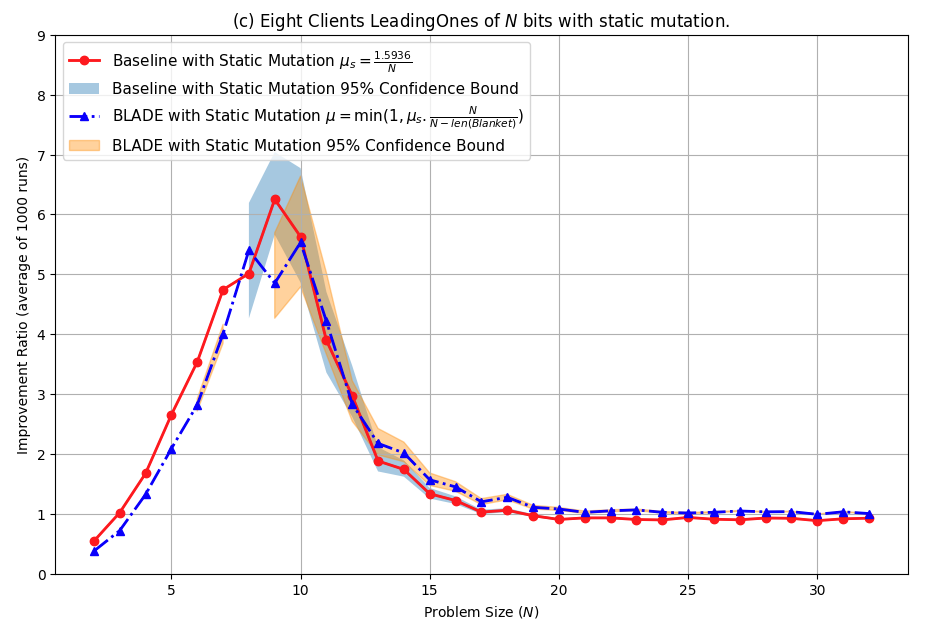}
  \includegraphics[width=0.48\linewidth]{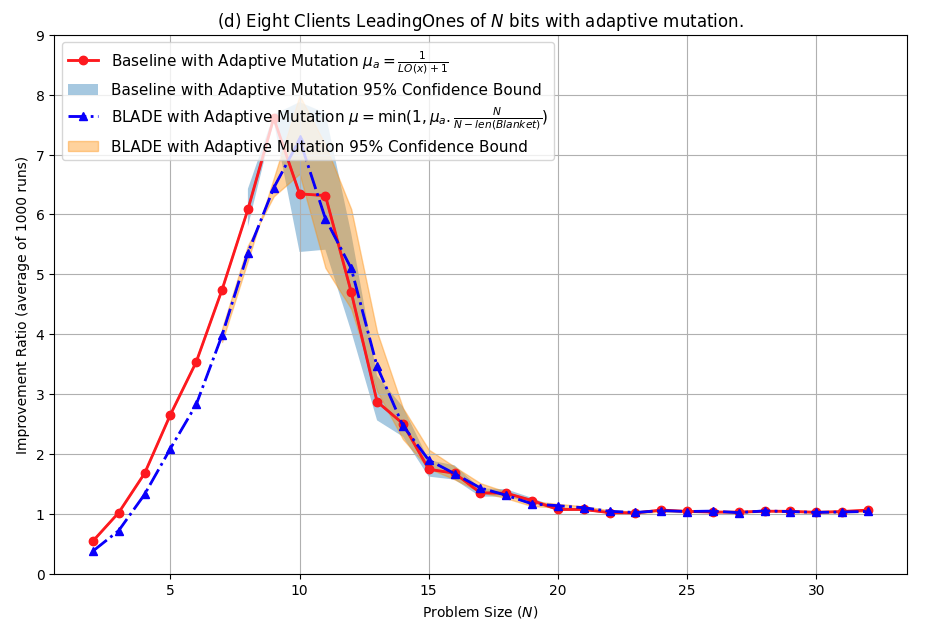}
  \caption{A comparison of the speedup ratio of BLADE vs.\ the baseline on the four benchmark problems with distribution over \emph{eight} clients;  the experimental and display details and conclusions are similar to those in Figure~\ref{fig:synergy}.}
  \Description{}
  \label{fig:A-2-3}
\end{figure*}

Figures~\ref{fig:A-2-1}, \ref{fig:A-2-2}, and~\ref{fig:A-2-3} compare the improvement ratios of BLADE and the baseline on the four benchmark problems with two, four, and eight clients, respectively.

\end{document}